\documentclass[letterpaper, 10 pt, conference]{ieeeconf}  

\IEEEoverridecommandlockouts                              

\usepackage{cite}
\usepackage{amsmath,amssymb,amsfonts}
\usepackage{algorithmic}
\usepackage{graphicx}
\usepackage{textcomp}
\usepackage{xcolor}
\usepackage{orcidlink}
\usepackage{subcaption}
\usepackage{booktabs}
\usepackage{multirow}
\usepackage{url}

\title{\LARGE \bf Hardware-triggered Time Synchronization of Roadside Multi-lidar, Multi-camera Measurement System for Accurate Data Alignment}

\author{%
\begin{tabular}{c}
Shiva Agrawal \orcidlink{0000-0001-8633-341X}, 
Savankumar Bhanderi \orcidlink{0000-0001-7257-6736}, 
Zhiran Yan \orcidlink{0009-0006-4091-2698}, 
and Gordon Elger \orcidlink{0000-0002-7643-7327} \\
\\
IIMo, Technische Hochschule Ingolstadt, Germany \\
{shiva.agrawal, savankumar.bhanderi, zhiran.yan, gordon.elger\}@thi.de}
\end{tabular}%
}

\begin{document}

\maketitle


\begin{abstract}
Accurate temporal alignment of heterogeneous sensors is necessary for reliable environment perception in roadside multi-lidar, multi-camera systems, particularly in dense urban traffic. For this purpose, an open-source, simple, modular, and configurable hardware-triggered time-synchronization circuit is presented in this work to perform temporal alignment or accurate time synchronization between a lidar and multiple cameras.  In the designed circuit, a lidar synchronization pulse is used as a reference input, and independently programmable, time-delayed trigger pulses are generated for each camera, allowing flexible adaptation to varying sensor setups and mounting geometries. A series of experiments is conducted on a roadside-mounted perception system comprised of lidar and three cameras, in which the trigger delay is systematically varied, and its impact on spatial-temporal alignment is evaluated. For different classes of road users, the overlap between lidar point cloud measurements and camera measurements is quantified to identify delay configurations that maximize cross-sensor consistency. The proposed circuit is shown to achieve robust and repeatable synchronization while remaining straightforward to deploy, reconfigure, and extend due to its simple and open-source design. Following validation on a three-camera roadside system, the circuit is extended to a vehicle platform with seven cameras and a lidar, providing a low-cost, extensible solution for multi-sensor synchronization across infrastructure and vehicle setups. All hardware circuit design files and source codes are available at \url{https://github.com/shiva-THI/hardware-trigger-time-sync-lidar-cameras}.
\end{abstract}

\section{Introduction} \label{sec:introduction}

In a multi-sensor system, each sensor generally operates independently, with its own clock and sampling rate. For example, cameras typically measure at rates like 30 Hz, lidar at 10 or 20 Hz, and radar at variable rates depending on the sensing mode and environment. However, to fuse the data from two or more sensors for accurate road user detection and motion prediction, the measurements from multiple sensors must be very well aligned temporally \cite{time_delay_measurement, time_sync_basics}. It implies that multi-sensors with overlapping fields of view must acquire data nearly simultaneously to capture the same physical world, and that each sensor's measurement data should then be timestamped as close as possible to the time of acquisition, ideally at the hardware level rather than later in the software pipeline, to minimize jitter and delay. 

Equally important is ensuring that all assigned timestamps are referenced to the same time base (common time reference), such as Coordinated Universal Time (UTC) or Global Positioning System (GPS) time (corrected to UTC), across sensors. Synchronizing clocks via high-precision protocols such as Precision Time Protocol (PTP) \cite{ieee_ptp} or Network Time Protocol (NTP) \cite{ieee_ntp} enables consistent, system-wide time labeling. Without a unified reference, even
accurately recorded local acquisition times on different sensors may drift, rendering data alignment unreliable and fusion errors more likely. Hence, three factors are important for overall accurate time-synchronization for multi-sensor measurements:

\begin{enumerate}
    \item Acquiring simultaneous measurements from multiple sensors with overlapping fields of view.
    \item Time stamping each data sample at the hardware level, very close to the instance when the sensor measures the physical environment, eliminating signal processing and network-induced timing uncertainty.
    \item Using a globally synchronized common time base for time stamping all the sensors.
\end{enumerate}

Hardware-based time synchronization \cite{mdpi_sustainability_mmwave_camera_fusion_review, CCIS_springer_chapter_2023} means that all the sensors are triggered at the hardware level to measure and send data simultaneously. This trigger is sent either by external hardware
or by one of the sensors of the system, acting as a master, and other sensors acting as slaves. Hardware triggering ensures that data acquisition begins at the same instance, eliminating timing offsets caused by asynchronous clocks or software delays. By time-stamping close to the actual measurement events in the early pipeline stages, the hardware-triggered time synchronization method significantly reduces jitter and accumulation of timing errors, resulting in highly accurate multi-sensor time synchronization.

However, hardware-triggered time synchronization is generally challenging and requires good knowledge of sensors in use and the ability to design custom electronics. Further, it is equally important that the delay for measurement is easily configurable and remotely accessible if the multi-sensor system is deployed on the road. While some prior works \cite{sommerLowcostSystemHighrate2017, osadcuksClockbasedTimeSynchronization2020, liuBriefIndustryPaper2021, wangHardwareBasedTimeSynchronization2024, gurumadaiahPreciseSynchronizationLiDAR2025}
explored external circuits for lidar-camera time synchronization, they lack configurable, remotely accessible delays for roadside deployments and open-source hardware designs for research reproducibility. 

Hence, to address these gaps, in the presented work, a roadside-mounted multi-sensor system comprising $2$ lidar sensors and $3$ cameras is synchronized temporally by designing an open-source, hardware-triggered time-synchronization electronic circuit. In the designed circuit, a lidar synchronization pulse is used as a reference input, and independently programmable, time-delayed trigger pulses are generated for each camera, allowing flexible adaptation to varying sensor positions. Furthermore, the designed electronic circuit is relatively simple and easily extendable to a larger number of cameras as required.

\subsection{Major contributions}
\begin{itemize}
    \item A roadside multi-lidar, multi-camera sensor system is developed for traffic monitoring and control.
    \item An open-source, hardware-triggered time-synchronization electronic circuit is designed and tested for a lidar and 3-camera setup with easily configurable time delays for each camera.
    \item The performance of the data alignment between lidar and camera is evaluated by systematically changing the time delay between measurements and calculating the coverage for different road users.
    \item The presented circuit is further extended to a lidar and 7-camera vehicle setup for accurate hardware-triggered time synchronization.
\end{itemize}

\section{State-of-the-art} \label{sec:sota}

Various strategies \cite{liGlobalClockSynchronization2004,nodaFrameSynchronizationNetworked2014,englishTriggerSyncTimeSynchronisation2015,hylaMultiCameraTriggering2016,sommerLowcostSystemHighrate2017,huSoftTimeSynchronization2018,osadcuksClockbasedTimeSynchronization2020,liuBriefIndustryPaper2021,grammatikopoulosEffectiveCameratoLidarSpatiotemporal2022,yuanLiCaS3SimpleLiDAR2022,wangTimeSynchronizationSpace2023,kuhseSyncSinkRobustness2024,wangHardwareBasedTimeSynchronization2024, gongRoadsideLiDARCameraFusion2025,gurumadaiahPreciseSynchronizationLiDAR2025, daiLiDARCameraSpatiotemporal2025} for multi-sensor time synchronization have been found in recent literature. However, among them, only \cite{sommerLowcostSystemHighrate2017, osadcuksClockbasedTimeSynchronization2020, liuBriefIndustryPaper2021, wangHardwareBasedTimeSynchronization2024, gurumadaiahPreciseSynchronizationLiDAR2025} have experimented with time synchronization of lidar and camera sensors using some form of dedicated hardware or external trigger methods, while other works have focused on software-based time synchronization, either using ROS or deep learning methods or some software algorithms. However, software-based time synchronization strategies are not as accurate as hardware-based strategies \cite{mdpi_sustainability_mmwave_camera_fusion_review, CCIS_springer_chapter_2023}.

From the above-stated hardware-based time synchronization work found in literature, authors in \cite{sommerLowcostSystemHighrate2017} used external photo diodes for laser detection and LED flashing for camera time stamping, but the system is not suitable for roadside installations where such external diodes and LEDs (light emitting diodes) are required to be installed in front of sensors at a height. Also, such time sync setups are vulnerable to external bad weather conditions. Authors in \cite{osadcuksClockbasedTimeSynchronization2020} used a PIC32 microcontroller to generate simultaneous trigger pulses to sync event camera, lidar, and ambient sensors; however, the details of the hardware design are not found. Authors in \cite{liuBriefIndustryPaper2021}  synchronized camera, lidar, radar, and IMU sensors using an FPGA-based SoC board, which is generally complex and costly to design and use. Additionally, the implementation details of this custom board are not provided by the authors. Further authors in \cite{wangHardwareBasedTimeSynchronization2024} designed a custom ARM Cortex-based board to synchronize camera, IMU, and lidar sensors at the hardware level, but the design is limited to a maximum of two cameras. Authors in \cite{gurumadaiahPreciseSynchronizationLiDAR2025} time synchronized one lidar and 3 cameras mounted on a test vehicle using a hardware-triggered approach. Authors also mentioned a delay offset estimation and correction hardware, but the details of this hardware are not found. 

From the literature review, it is evident that the work in the hardware-based time synchronization context for lidar and camera sensors is very limited, and further extensive details for designing a hardware-trigger circuit are not found that can be easily reused by the research community. Hence, the main objective of this paper is to provide a simple, open-source, modular, and easily configurable hardware-trigger circuit for accurately synchronizing lidar and camera sensors for environmental perception. The designed circuit is deployed on the roadside-mounted setup and tested through real-time experiments in a continuously working system.

\section{Roadside Multi-sensor Measurement System} \label{sec:measurement_setup}

\begin{figure}[!h]
\centering
\includegraphics[width=0.48\textwidth]{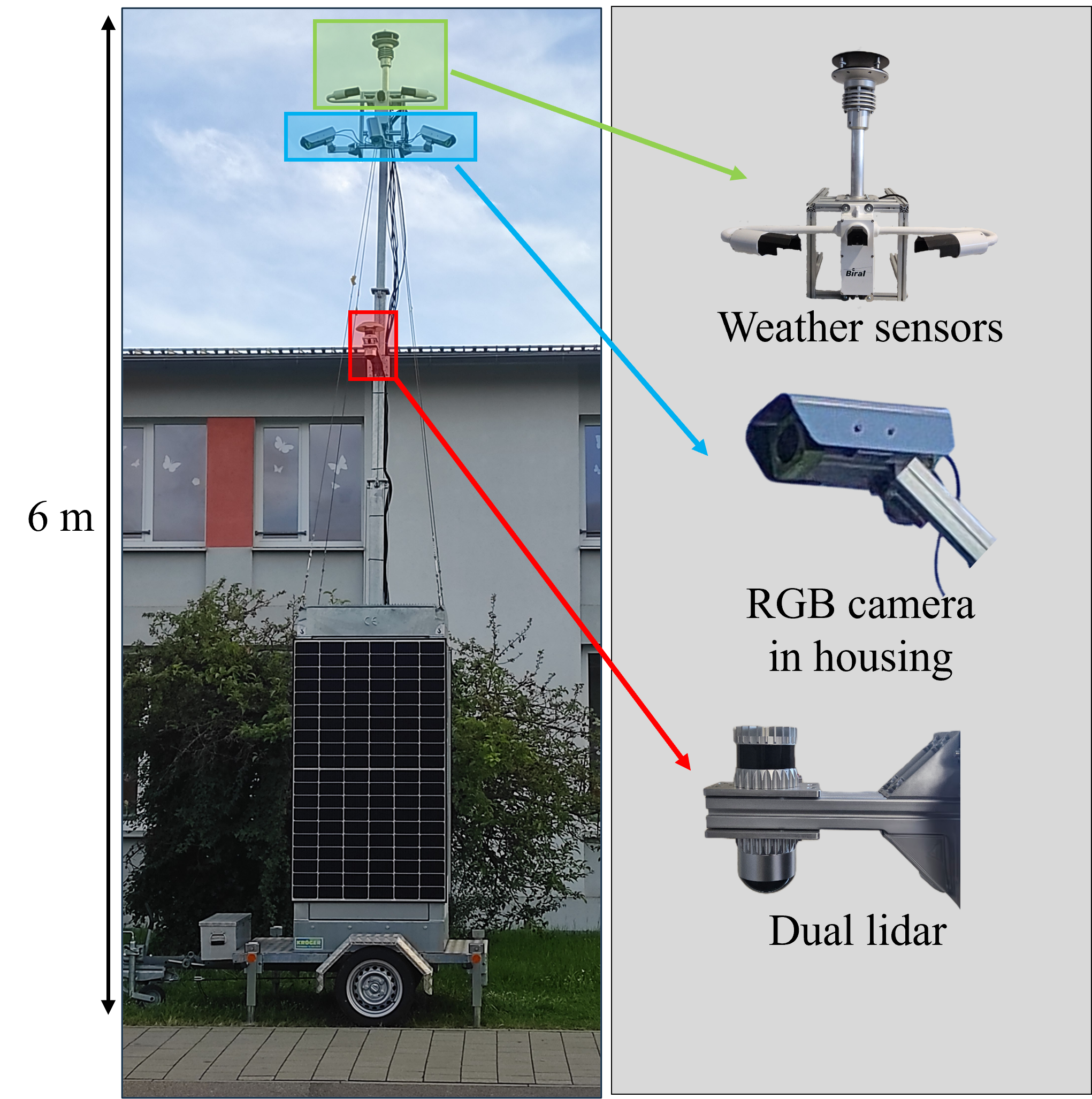}
\caption{Multi-sensor measurement system on a mobile mast. Weather sensors are highlighted in green, three RGB cameras in blue, and two lidar sensors in red.} 
\label{fig:inani_mobile_mast}
\end{figure}

The roadside-mounted multi-sensor measurement setup is comprised of two lidar sensors, three RGB cameras, and two weather sensors, as shown in Fig. \ref{fig:inani_mobile_mast}. The complete setup is deployed using a mobile mast for ease of transportation. The maximum height of the setup is $6$ m from the ground, and it is installed near the elementary school in front of the traffic light crossing. The two lidar sensors are Ouster OS1 \cite{ousterOS1_datasheet} with $64$ channels (mounted upright) and Ouster OSdome \cite{ouster_osdome_datasheet} with $128$ channels (mounted inverted below the OS1 sensor). Both lidar sensors are mounted at $4.5$ m height and arranged such that there is no blind spot near the pedestrian crossing in front of the sensor mast. The three RGB cameras are GigE Vision Basler ace2 a2a1920-51gcBAS \cite{basler_gige_camera} cameras, featuring an Ethernet interface that powers the sensors through Power-over-Ethernet and transmits the camera data to the processing device. Each camera also has a dedicated GPIO pin to send the external trigger signal for hardware-based time synchronization. All three cameras are mounted at an approximate height of $5.5$ m from the ground and are oriented to cover the entire street in front of the school with minimal overlapping field-of-view. The two weather sensors are mounted on the top of the sensor mast above the RGB cameras for weather-based data collection. However, the weather sensors are out of scope of the presented work on hardware-based time synchronization. 

\begin{figure}[!h]
\centering
\includegraphics[width=0.48\textwidth]{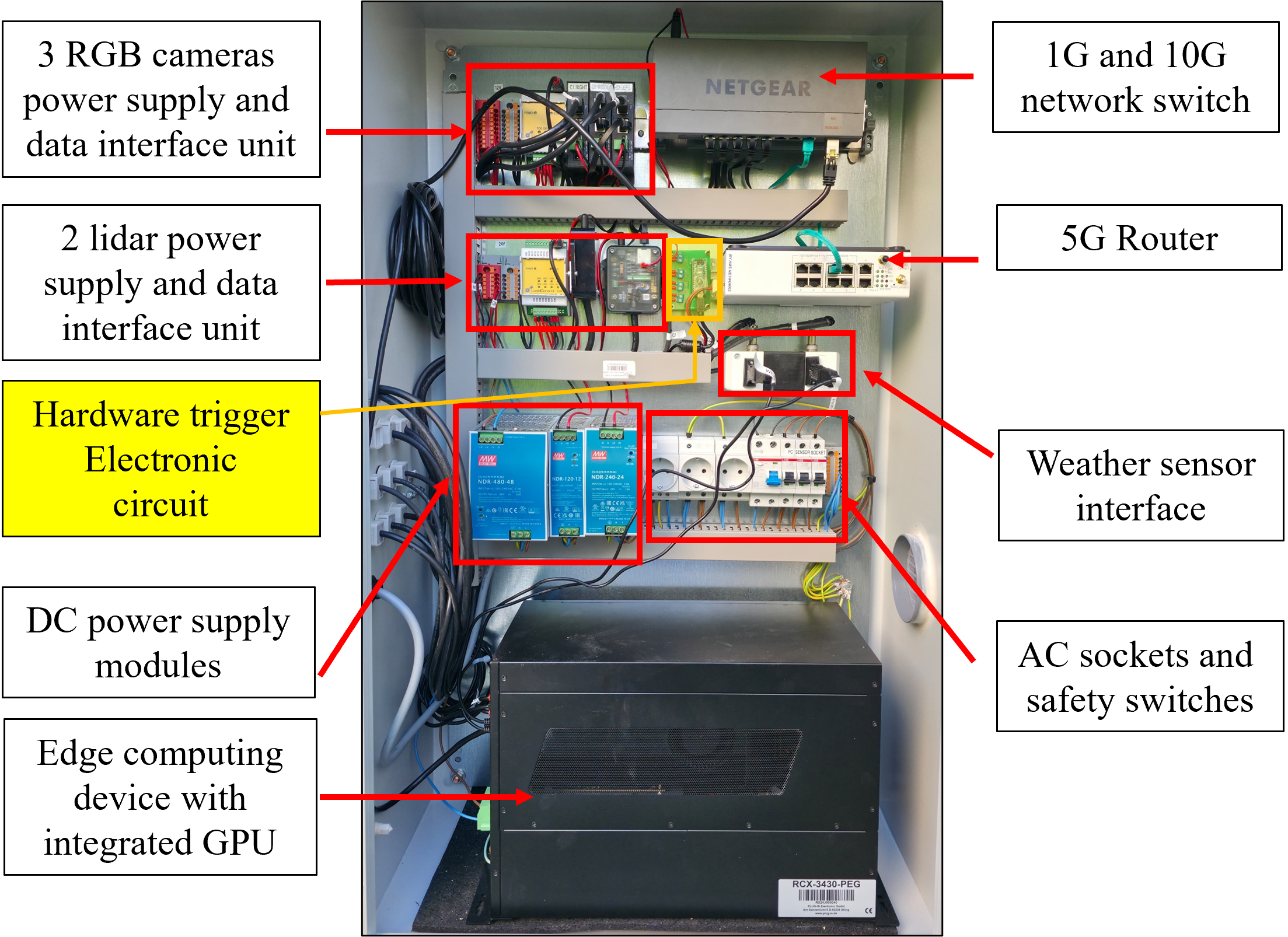}
\caption{Control cabinet of the deployed multi-sensor measurement system with highlighted details of components. The component shown in yellow is the electronic circuit for hardware triggering, synchronizing the lidar and three cameras temporally.} 
\label{fig:inani_control_cabinet}
\end{figure}

Fig. \ref{fig:inani_control_cabinet} shows the control cabinet of the deployed multi-sensor measurement system with highlighted details of individual components or modules. This cabinet is installed inside the mobile mast of Fig. \ref{fig:inani_mobile_mast}. The component shown in yellow is the electronic circuit for hardware triggering, synchronizing the lidar and three cameras temporally.

For accurate data alignment of the dual lidar (two lidar setup) and three cameras, both spatial calibration and time synchronization are required. For spatial calibration of the sensors, the multi-sensor calibration methodology of \cite{shiva_auto_calibration_paper} is utilized. This methodology uses static calibration targets to calibrate camera, radar, and lidar sensors. However, in this setup, it is used to calibrate two lidars and three cameras. For the complete setup, the lidar OS1 mounted upright is considered as the reference sensor, and hence all other sensors are calibrated with respect to this sensor. Once the spatial calibration is performed and validated using the static environment within the sensor field-of-view, time synchronization is performed between each camera and the lidar OS1 using the designed hardware trigger circuit as explained in next section.

\section{Hardware trigger time synchronization of lidar and cameras} \label{sec:hardware_trigger}

As stated before, the Ouster OS1 lidar sensor, which is used as the reference sensor for the complete setup, is also used as the master sensor to control the trigger of all three cameras. The OSDome lidar does not generate trigger pulses for the cameras but requires precise synchronization with the OS1 lidar for the overall working of the multi-sensor system. However, ouster provides built-in phase-lock functionality for multi-lidar synchronization, enabling configurable angular offsets between sensors. Hence, to synchronize both lidar sensors with each other, this built-in phase-lock feature is enabled in their respective configurations with a phase angle of $0^\circ$. This approach achieves precise time synchronization, aligning the start of both lidar rotations to multiples of $100$ ms (e.g., $100$, $200$, $300$, $400$, $500$ ms) at $0^\circ$ when operating both sensors at $10$ Hz. Further, both lidar sensors are synchronized with the system time from the edge computing device (refer to Fig. \ref{fig:inani_control_cabinet}) as a common reference time using PTP \cite{ieee_ptp} to generate hardware-level high-accuracy time stamps very close to the time of measurement. The edge computing device maintains UTC synchronization (adjusted to the local time) via the internet connection. This precise timestamp is then available directly in the header of the ROS 2 point cloud message \cite{ros_pc2_msg} of each lidar measurement frame, which is available from the lidar ROS 2 driver. In this work, ROS 2 Humble \cite{ros2} with Ubuntu 22.04 LTS is used for multi-sensor data processing.

\begin{figure*}[!h]
\centering
\includegraphics[width=0.60\textwidth]{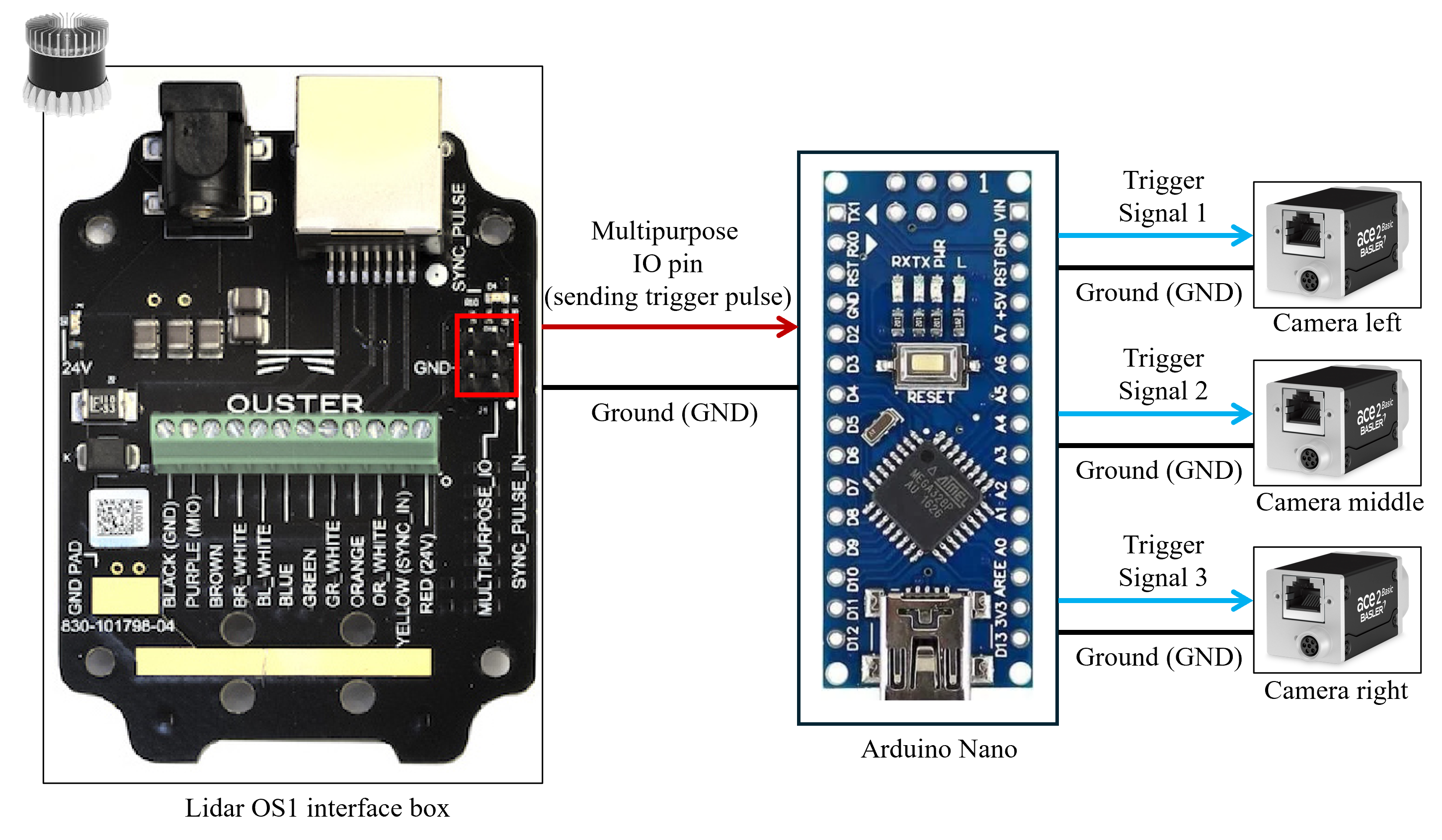}
\caption{Hardware trigger high-level architecture. On the left is the lidar interface box \cite{ouster_interface_box}, in the centre is the Arduino Nano v3, and on the right are the Basler cameras \cite{basler_gige_camera}. Signal lines (shown in red and cyan color) indicate direction of trigger signal flow, and GND lines provide a common reference}
\label{fig:trigger_architecture}
\end{figure*}

Ouster OS1 lidar has a multipurpose IO (input/output) pin provided by the manufacturer through the interface box of the sensor (please refer to section $7$ of the hardware user manual \cite{ouster_os1_hardware_user_manual_rev7} and Fig. \ref{fig:trigger_architecture} (left)). This pin is configurable to either receive a PPS (pulse per second) signal from an external GPS sensor to synchronize the lidar sensor timestamp with GPS time, or it can be configured as an output pin to send the trigger pulse for hardware triggering other sensors. In the presented multi-sensor setup, this pin is used to generate trigger pulses configured to emit one pulse at the start of each scan cycle. This results in $10$ Hz pulses matching the lidar rotation frequency, precisely aligned to multiples of $100$ ms. The main configuration parameters used to generate lidar trigger pulses are as follows:
\begin{itemize}
    \item Lidar Mode: $1024 \times 10$
    \item Timestamp Mode: Time from PTP 1588
    \item Multipurpose IO Mode: Output from encoder angle
    \item Phase lock: Enable
    \item Phase lock Polarity: $0^\circ$
\end{itemize}

\begin{figure*}[!h]
\centering
\includegraphics[width=\textwidth]{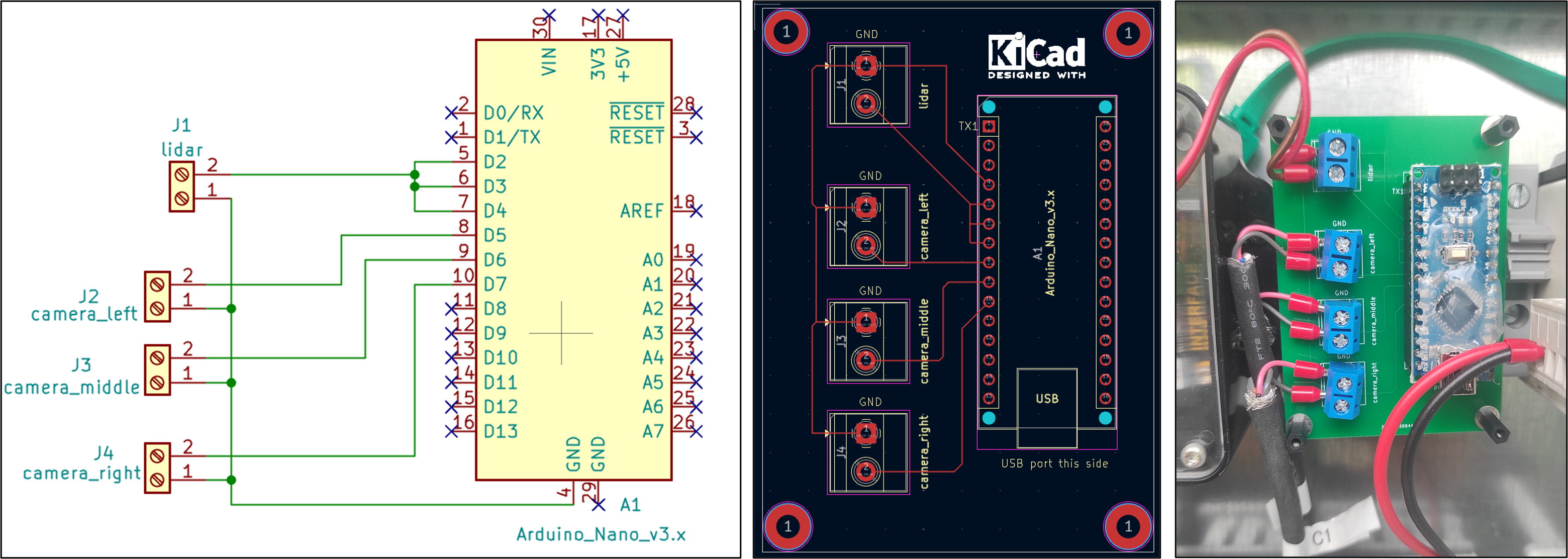}
\caption{Hardware trigger circuit for time synchronization of lidar with three cameras: (left) schematic design, (center) KiCad-generated PCB layout, and (right) deployed PCB in control cabinet (USB power cable omitted).} 
\label{fig:three_camera_complete_design}
\end{figure*}

The multi-purpose IO pin in the lidar OS1 interface box is connected to the VCC (DC power supply) through the available pull-up resistor via a Jumper. This provides more current to the trigger pulse as compared to the default pull-down mode. However, the amount of current available after pull-up is sufficient to directly trigger only up to three cameras (in this case, Basler ace2 a2a1920-51gcBAS) reaching the maximum current limit. Hence, to develop a generic solution that can be extended to many cameras, an intermediate Arduino Nano is used as an intelligent and configurable device that receives the lidar trigger pulse as input during every scan and generates individual output trigger pulses for cameras in that scan. By introducing the open-source and low-cost Arduino Nano, the hardware trigger circuit is easily expandable to many more cameras with different delay configurations as per the required mounting settings of sensors on the road. Furthermore, the Arduino Nano is connected to the Edge computing device of the sensor setup (refer to Fig. \ref{fig:inani_control_cabinet}) via the USB cable. This connection provides a power supply, optional serial communication, and the ability to update the delay time of each camera remotely for experiments.

Fig. \ref{fig:trigger_architecture} illustrates the high-level hardware trigger architecture, while Fig. \ref{fig:three_camera_complete_design} presents the three-camera trigger circuit schematic (left), corresponding PCB layout (center), and fabricated board deployed in the roadside sensor setup (right). In the schematic diagram, the multi-purpose IO pin of lidar OS1 that generates a trigger pulse is connected to pin-2 of connector J1, and the GND pin of the lidar interface box is connected to pin-1 of connector J1. Similarly, the trigger input cable of each camera is connected directly to pin-2 of connectors J2, J3, and J4, while their respective ground cables are connected to pin-1 of connectors J2, J3, and J4. On the Arduino side, the lidar trigger pulse pin is then connected to digital input pins D2, D3, and D4 to generate individual and independent interrupt routines for each camera. Similarly, the trigger pins of the camera are connected to digital output pins D5, D6, and D7 of Arduino for the left camera, middle camera, and right camera, respectively. The ground connections from all four connectors are given to the Arduino ground pin. 

While the Arduino Nano's interrupt routine jitter may introduce minor temporal variations (typically in a few microseconds), it does not cause cumulative drift because the system re-aligns to the lidar’s phase-locked reference every cycle. This jitter may be considered negligible as it remains orders of magnitude lower than the non-deterministic latencies found in software-based synchronization.

Hence, for three cameras, three IO (input-output) pairs are formed to define dedicated control for each camera. These are pin-pairs D2-D5, D3-D6, and D4-D7. For example, pin-pair D2-D5 is for controlling the trigger signal of the left camera, where pin D2 acts as an input trigger obtained from lidar and pin D5 acts as an output to send a trigger pulse with the required delay to the left camera. This pin-pair design facilitates efficient debugging and verification of circuit performance, while enabling modular expansion or reduction of camera channels as needed. Additionally, this pin-pair design supports multi-lidar, multi-camera configurations, where each lidar can independently trigger dedicated camera subsets, if required. As an example, a configuration with three lidar sensors, each synchronizing three cameras (say), is readily achieved by instantiating dedicated lidar input and camera output pin-pairs within this versatile architecture.

Analogous to the lidar sensors, the ROS 2 camera driver processes data from each camera after enabling PTP settings, chunk time configurations, and hardware trigger parameters. Timestamps precisely aligned to the hardware delays, derived from the PTP-synchronized edge computing device clock, are incorporated into the ROS 2 Image message headers \cite{ros_image_msg} of each camera measurement frame, ensuring temporal consistency. It should be emphasized that, despite using ROS 2 for data collection and processing, the multi-sensor system's time synchronization is implemented entirely via hardware triggering, with no software-based temporal adjustments.  

Additionally, it is important to note that in the multi-sensor setup described before in Section \ref{sec:measurement_setup}, when lidar starts scanning, it first covers the field of view of left camera, then middle camera, and then right camera. Hence, the time delay required to align each camera measurement with the scanning lidar measurement varies as per their relative positions on the setup. However, the exact relative position of the overlapping field of view between each camera and lidar is not measurable directly, and hence systematic time delays are introduced between each camera and lidar, and then experiments are performed to find the appropriate time delay to achieve the closest data alignment consistency. The next section describes the details of the performed experiments with results and discusses them accordingly.

\section{Experimental results and discussions} \label{sec:experimental_results_and_discussions}

To systematically assess the data alignment consistency achieved through hardware-triggered time synchronization between the three cameras and the dual-lidar system, multiple measurements are conducted under varying time delay configurations. Three distinct dynamic road users, a car, a bicycle, and a person or pedestrian, are included in the evaluation. The car is driven at a constant speed of $30$ km/h, corresponding to the maximum speed limit in the school zone, while the bicycle and pedestrian move at their respective normal speeds. Data is recorded across several short scenes, each configured with different time offsets between the lidar and camera systems, where every road user traverses the full field of view of the sensor setup in both directions. 

\begin{figure}[!ht]
\centering
\includegraphics[width=0.48\textwidth]{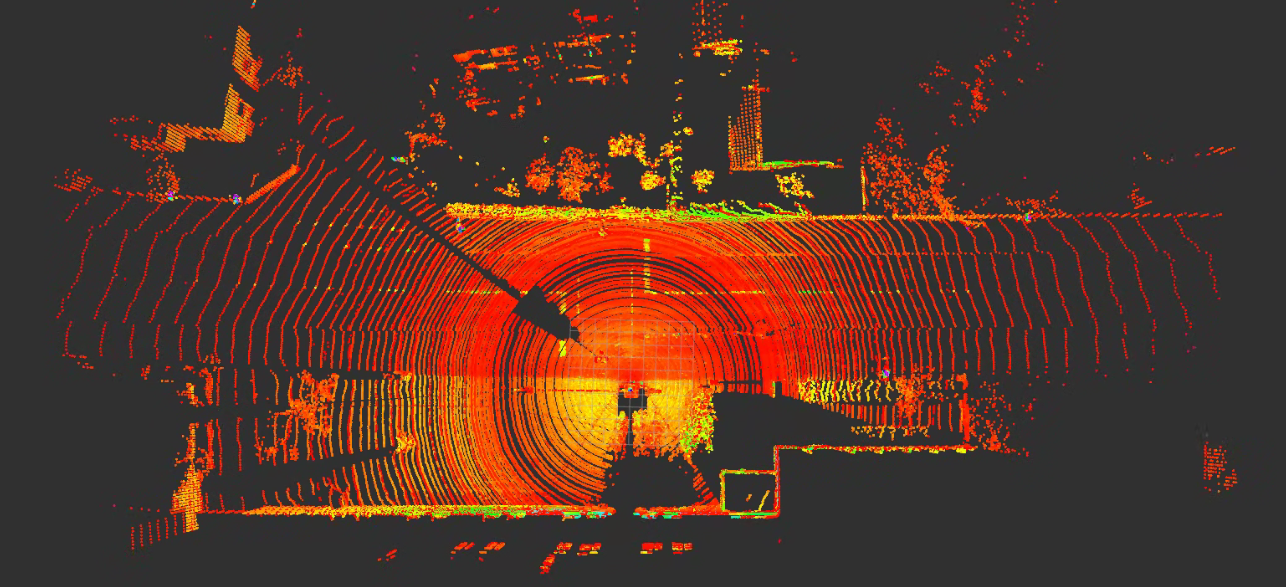}
\caption{One frame of the time-synchronized and merged point cloud of two lidar sensors on the described roadside setup.} 
\label{fig:dual_lidar_point_cloud_sample}
\end{figure}

During the evaluation of each scene, at first, the point cloud of two lidar sensors is merged into one point cloud for every frame using the spatial calibration and phase-locked time synchronization. One sample output of the time-synchronized and merged point cloud of the dual lidar is shown in Fig. \ref{fig:dual_lidar_point_cloud_sample} for reference.

For evaluating each time-delay configuration for each camera, images are first selected with the required road user in the field of view of the respective camera from near to far distances. As the multi-sensor setup is installed in front of the school, very close to the traffic light crossing for enhancing the safety of the children near school, the near distance area from the sensors, i.e., from $0$ to $15$\,m in both directions of the sensor setup, plays a much vital role. Thereafter, the mid range is defined from $15$\,m to $30$\,m, and the far range is defined from $30$\,m to $50$\,m. Beyond $50$\,m, the lidar point density reduces drastically, especially for persons and bicycles, and hence, the evaluations are performed till this distance.

After selecting multiple images from each camera, an object mask is generated using state-of-the-art Mask-RCNN (Mask Region-based Convolutional Neural Network) \cite{mask_rcnn} for each road user under evaluation. During this step, all other road users are not considered. Similarly, the merged lidar frames are labeled manually (using sustechpoints open-source tool \cite{sushtech_points_labeling_tool}) by annotating 3D bounding boxes to find the lidar points that belong to each of the three road users in each frame. Then, during evaluation, the point cloud belonging to each road user is projected on the respective image mask, and then the percentage of coverage is calculated for each frame as per \eqref{eq_per_coverage}. Further, the average coverage is calculated separately using all the frames within the given distance range.

\begin{equation} \label{eq_per_coverage}
\%\ \text{coverage} = 
\left( \frac{\text{projected lidar points inside mask}}{\text{total lidar points}} \right) \times 100
\end{equation}

\begin{figure*}[!h]
  \centering
  \begin{subfigure}[b]{0.32\textwidth}
    \centering
    \frame{\includegraphics[width=\textwidth]{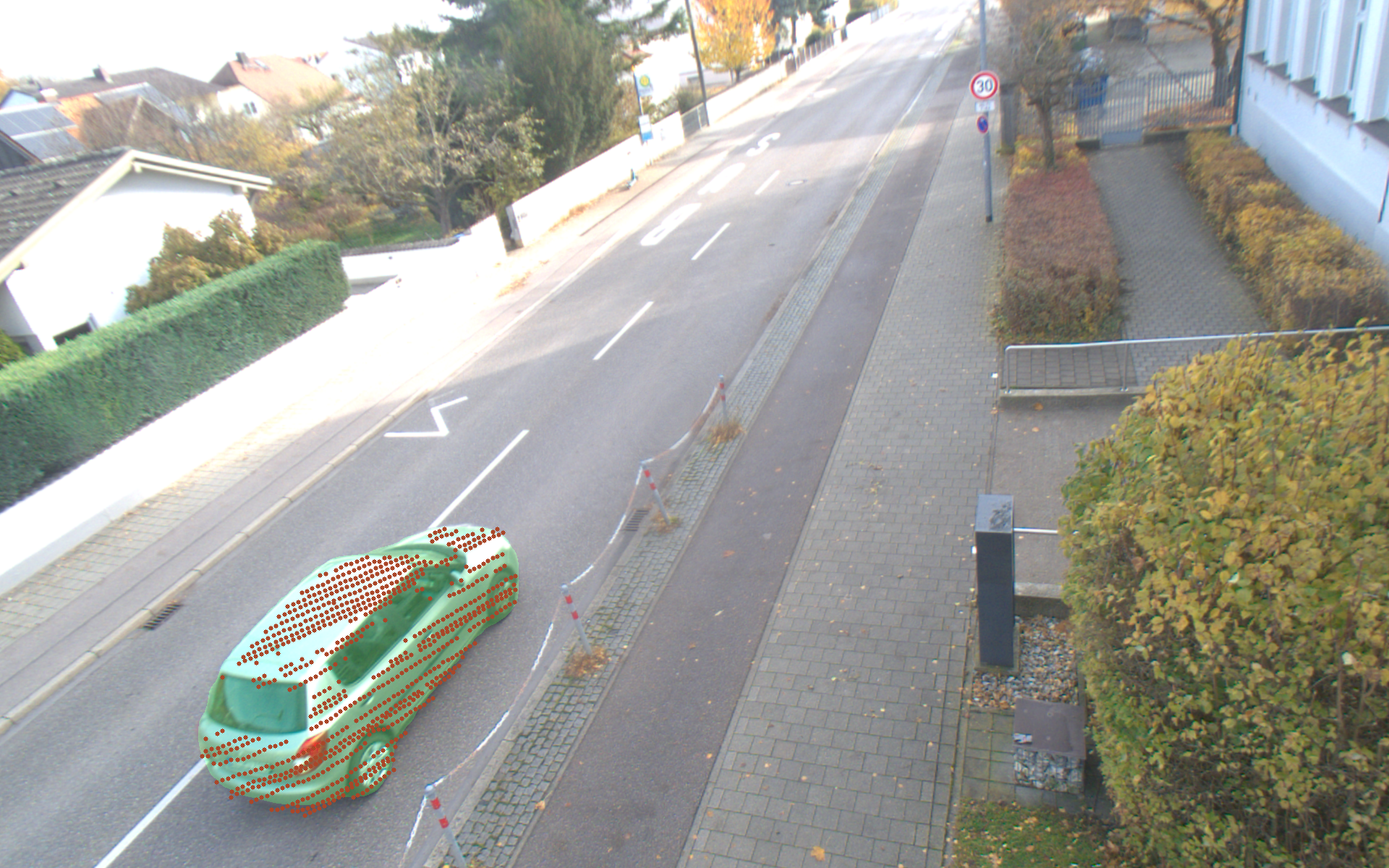}}
    \caption{Car in left camera}
    \label{subfig:camera_1_car}
  \end{subfigure}
  \begin{subfigure}[b]{0.32\textwidth}
    \centering
    \frame{\includegraphics[width=\textwidth]{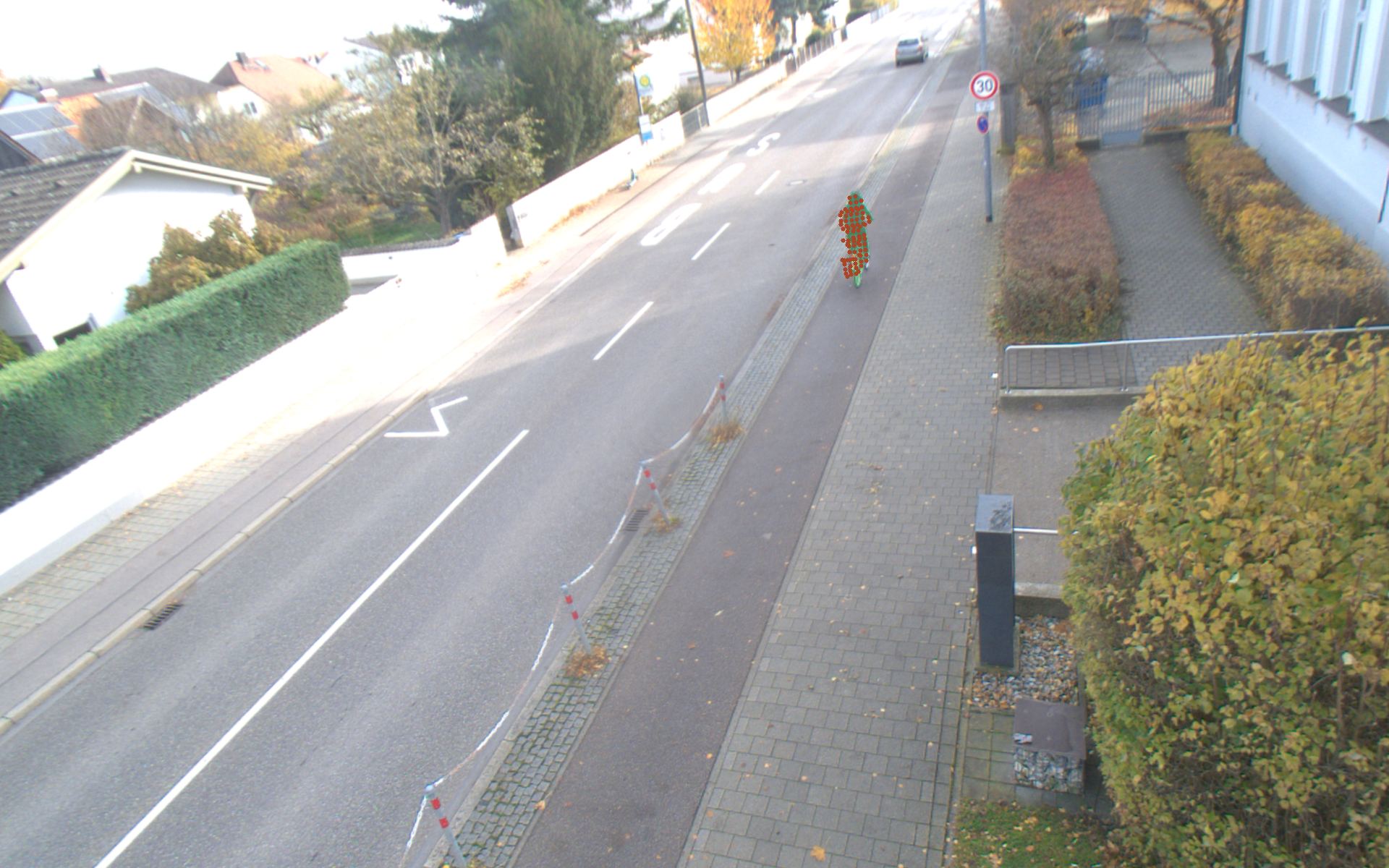}}
    \caption{Bicycle in left camera}
    \label{subfig:camera_1_bicycle}
  \end{subfigure}
  \begin{subfigure}[b]{0.32\textwidth}
    \centering
    \frame{\includegraphics[width=\textwidth]{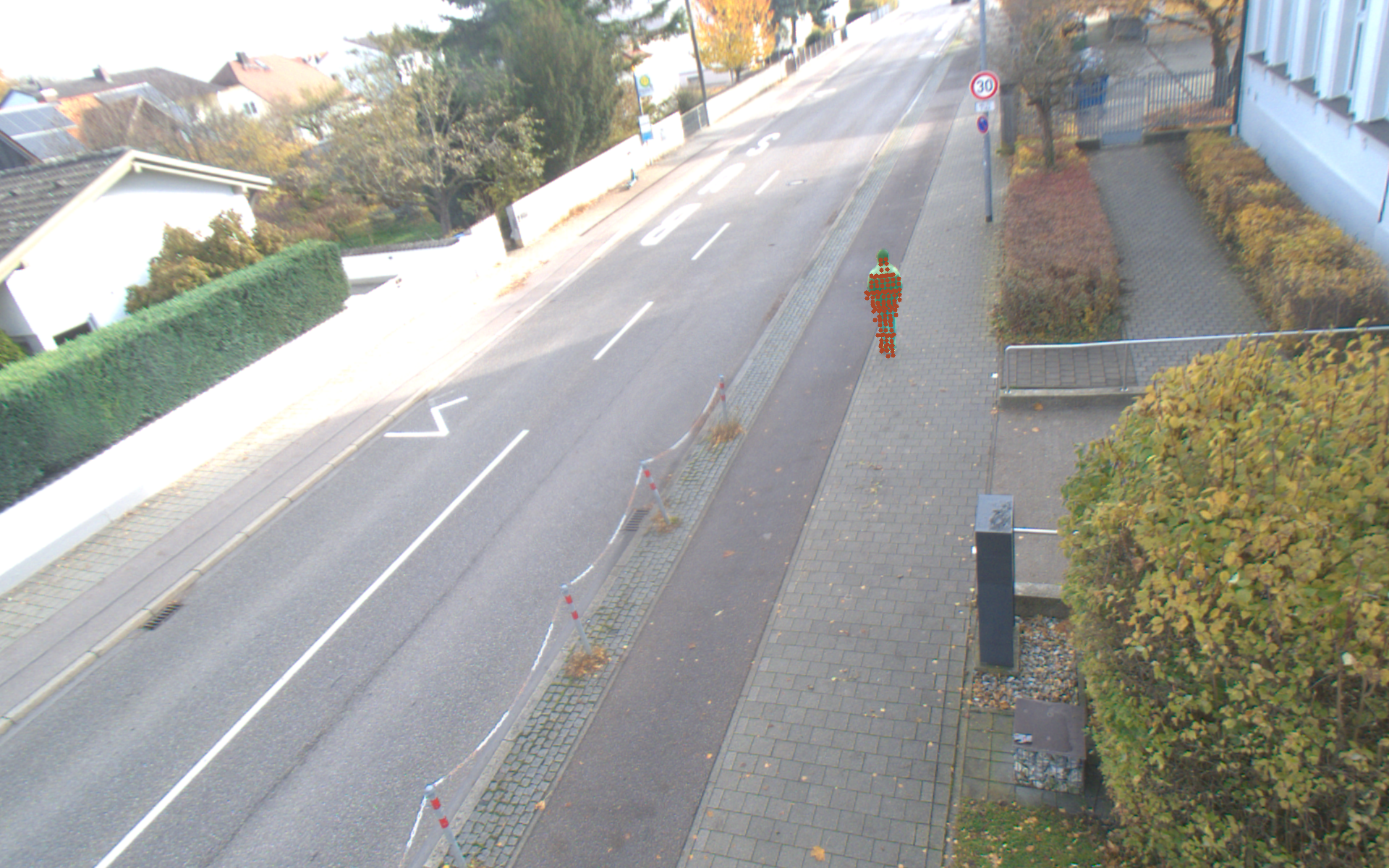}}
    \caption{Person in left camera}
    \label{subfig:camera_1_person}
  \end{subfigure}

  \vspace{0.1cm}

    \begin{subfigure}[b]{0.32\textwidth}
    \centering
    \frame{\includegraphics[width=\textwidth]{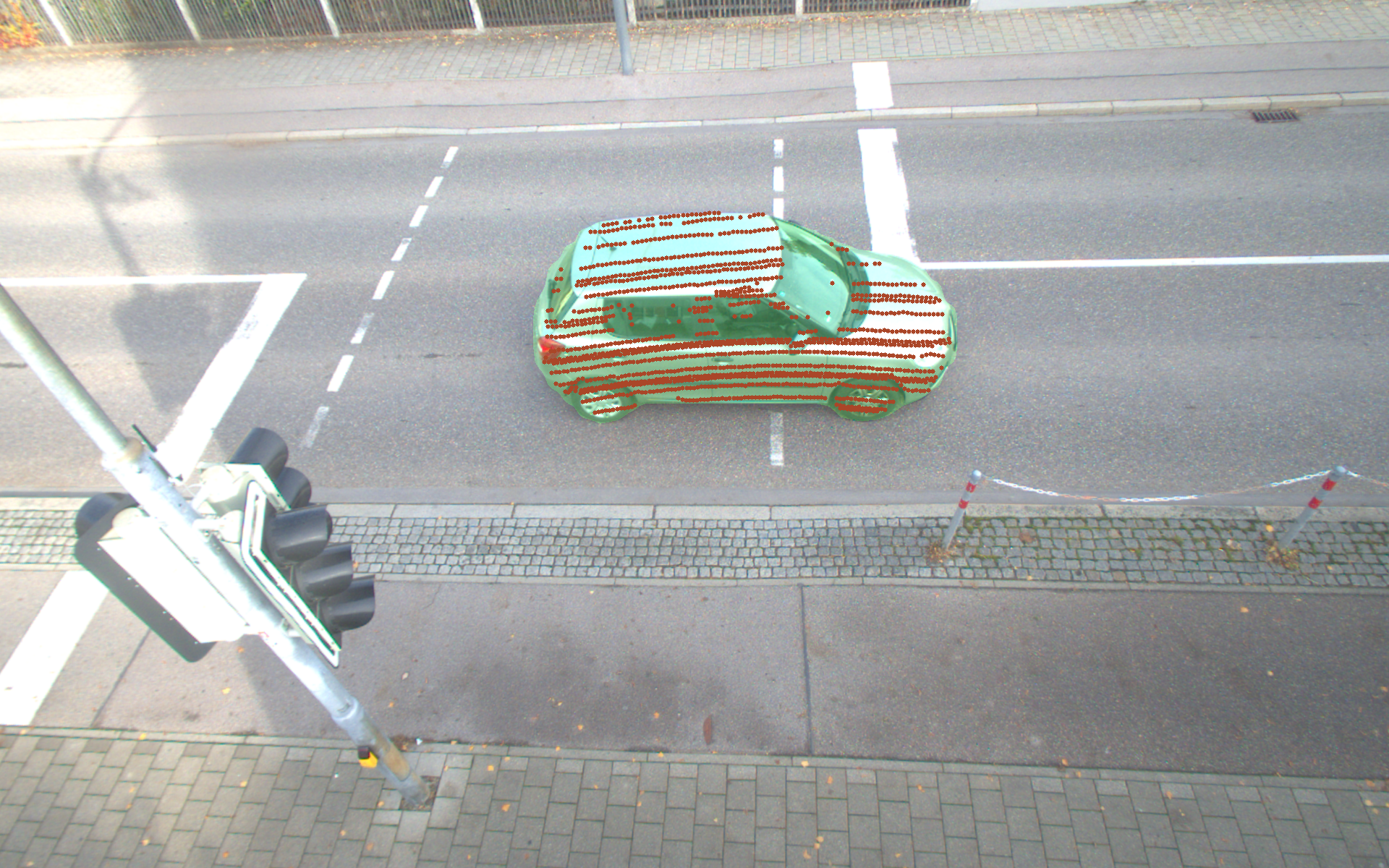}}
    \caption{Car in middle camera}
    \label{subfig:camera_2_car}
  \end{subfigure}
  \begin{subfigure}[b]{0.32\textwidth}
    \centering
    \frame{\includegraphics[width=\textwidth]{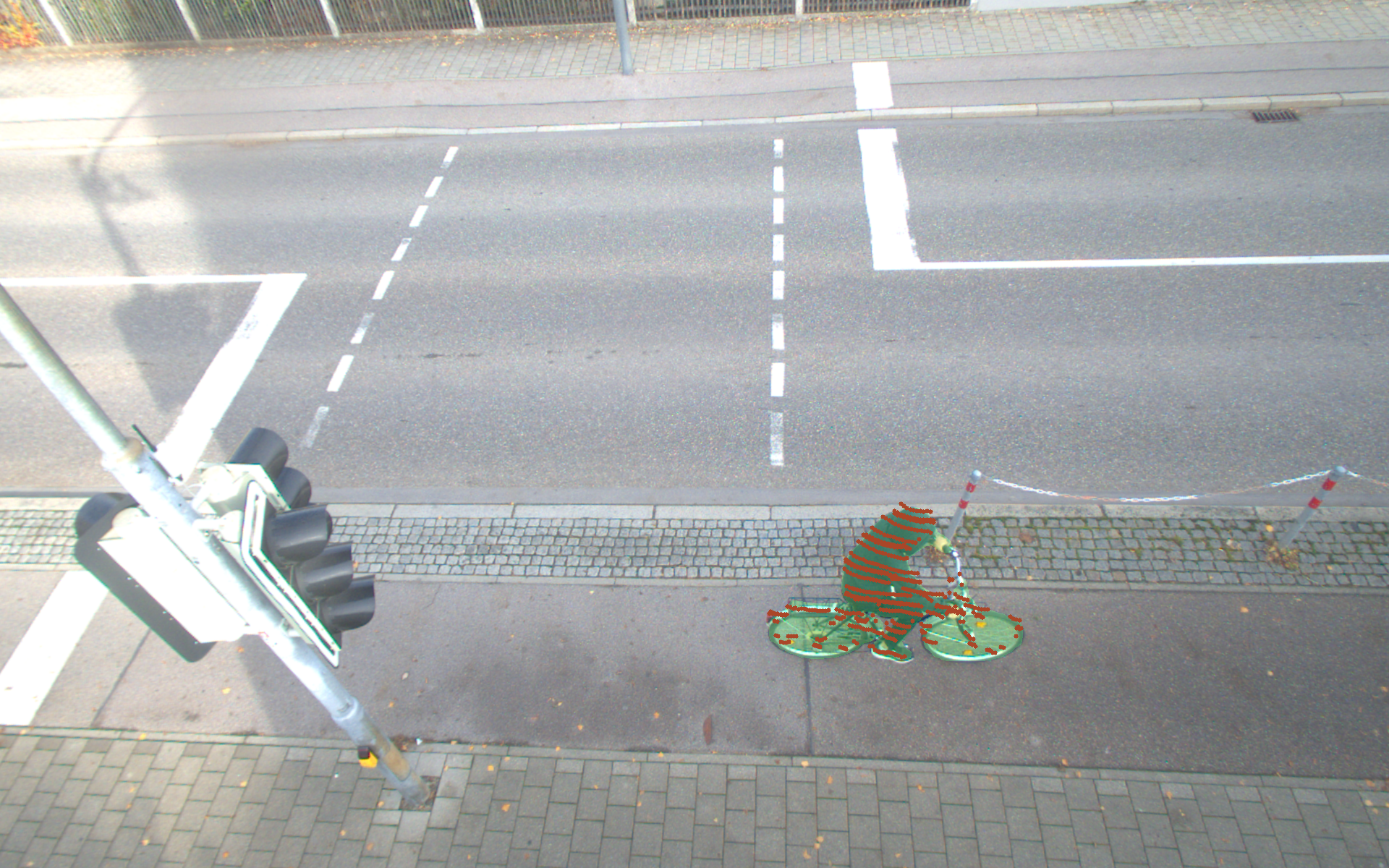}}
    \caption{Bicycle in middle camera}
    \label{subfig:camera_2_bicycle}
  \end{subfigure}
  \begin{subfigure}[b]{0.32\textwidth}
    \centering
    \frame{\includegraphics[width=\textwidth]{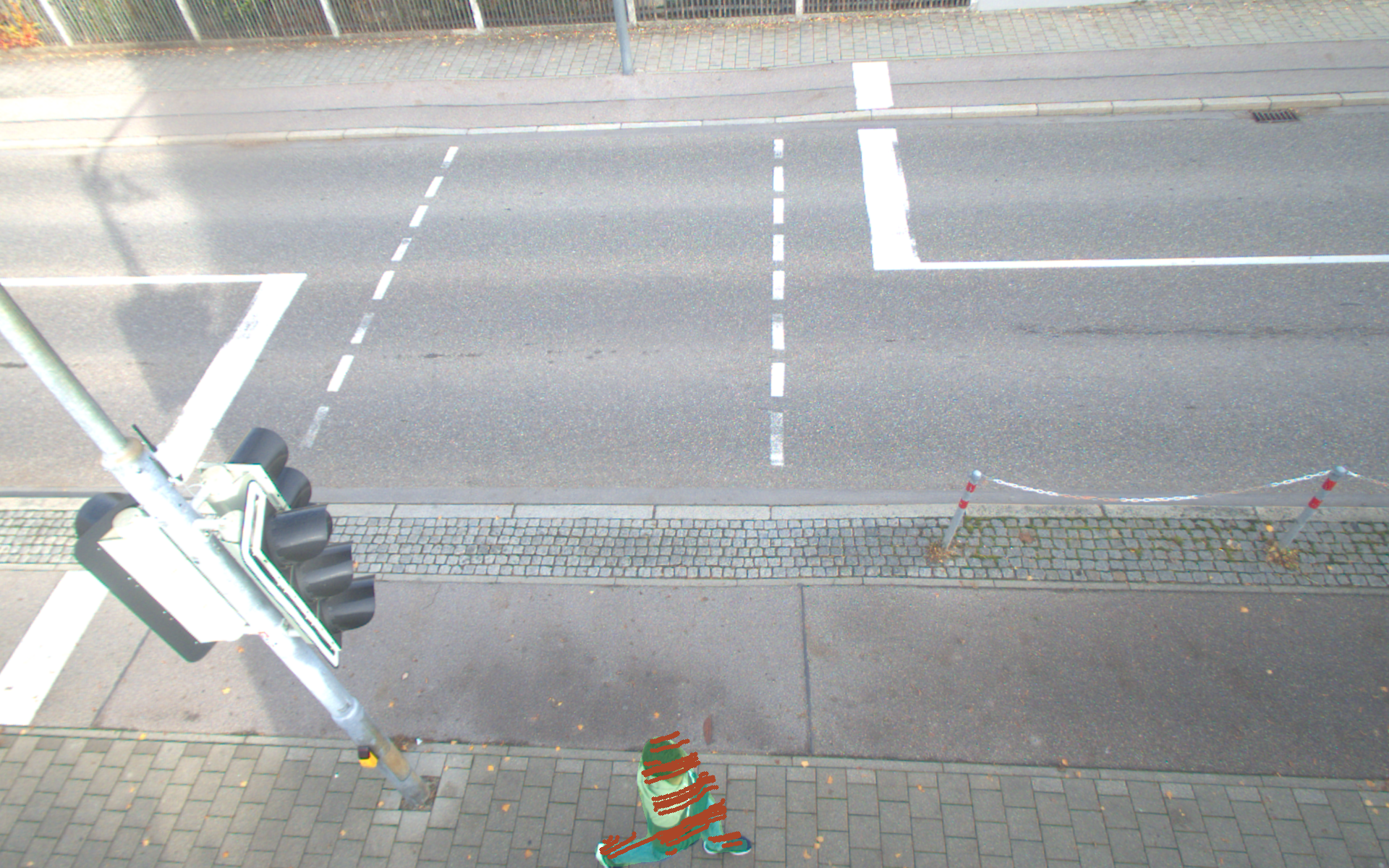}}
    \caption{Person in middle camera}
    \label{subfig:camera_2_person}
  \end{subfigure}

  \vspace{0.1cm}

    \begin{subfigure}[b]{0.32\textwidth}
    \centering
    \frame{\includegraphics[width=\textwidth]{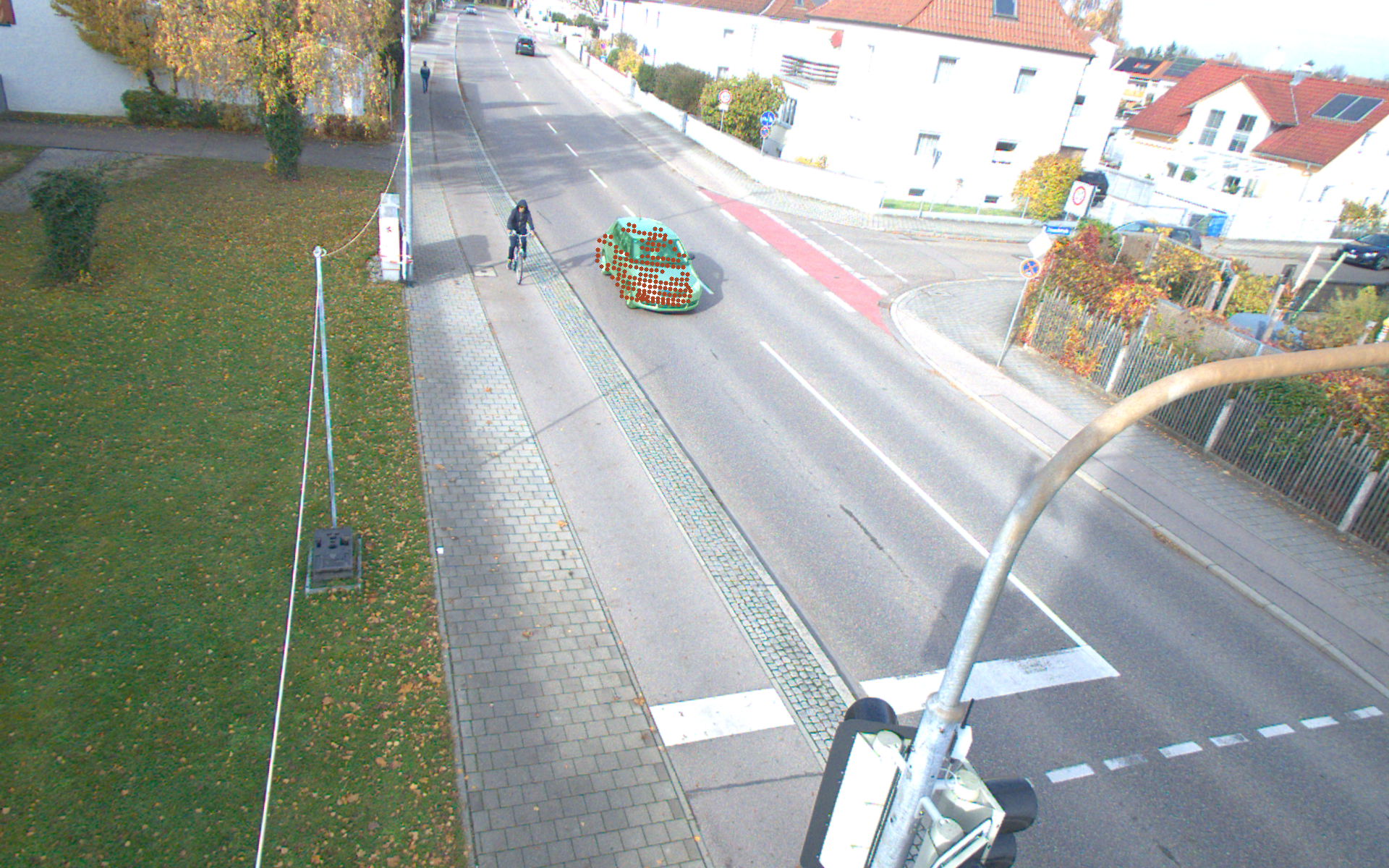}}
    \caption{Car in right camera}
    \label{subfig:camera_3_car}
  \end{subfigure}
  \begin{subfigure}[b]{0.32\textwidth}
    \centering
    \frame{\includegraphics[width=\textwidth]{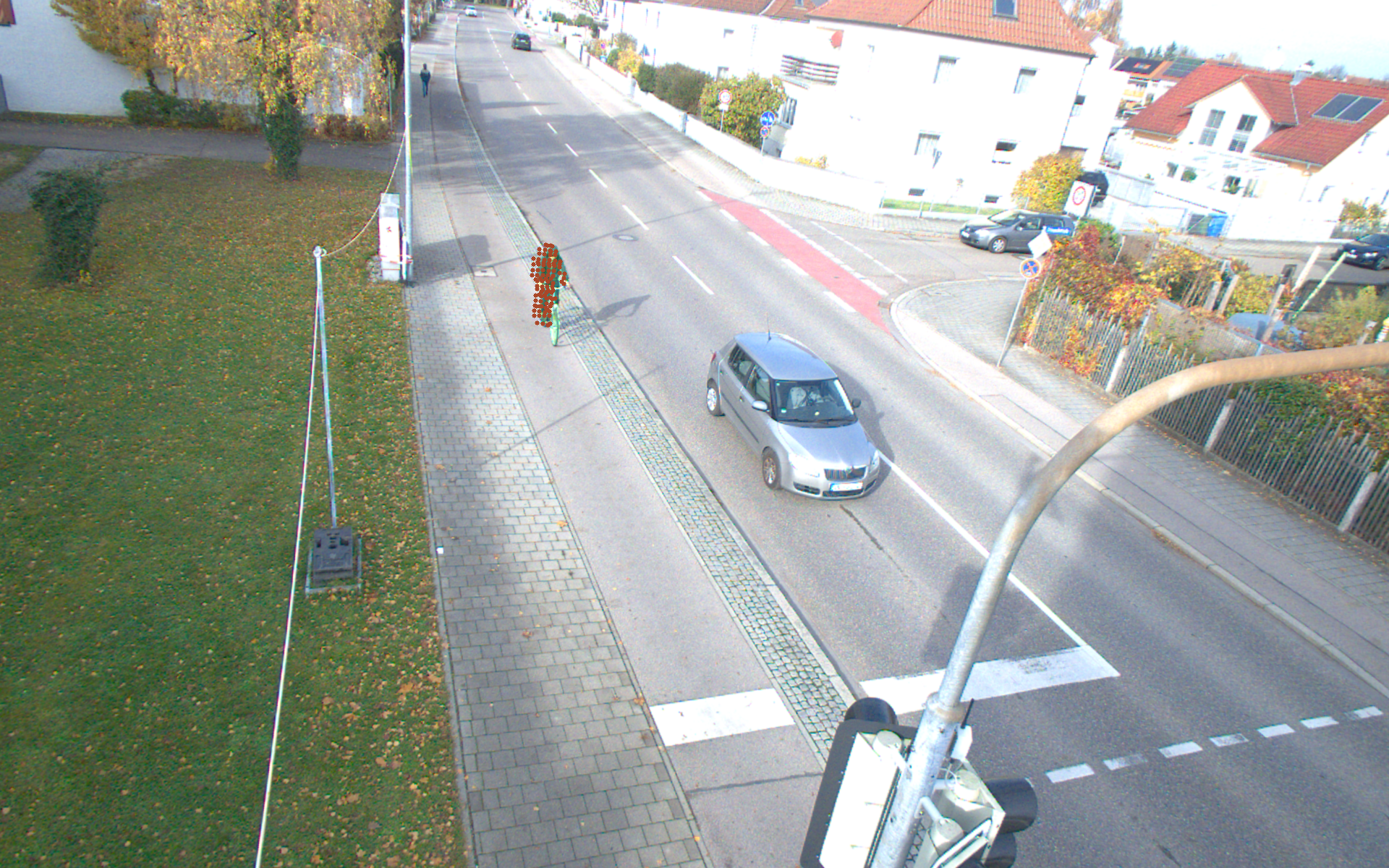}}
    \caption{Bicycle in right camera}
    \label{subfig:camera_3_bicycle}
  \end{subfigure}
  \begin{subfigure}[b]{0.32\textwidth}
    \centering
    \frame{\includegraphics[width=\textwidth]{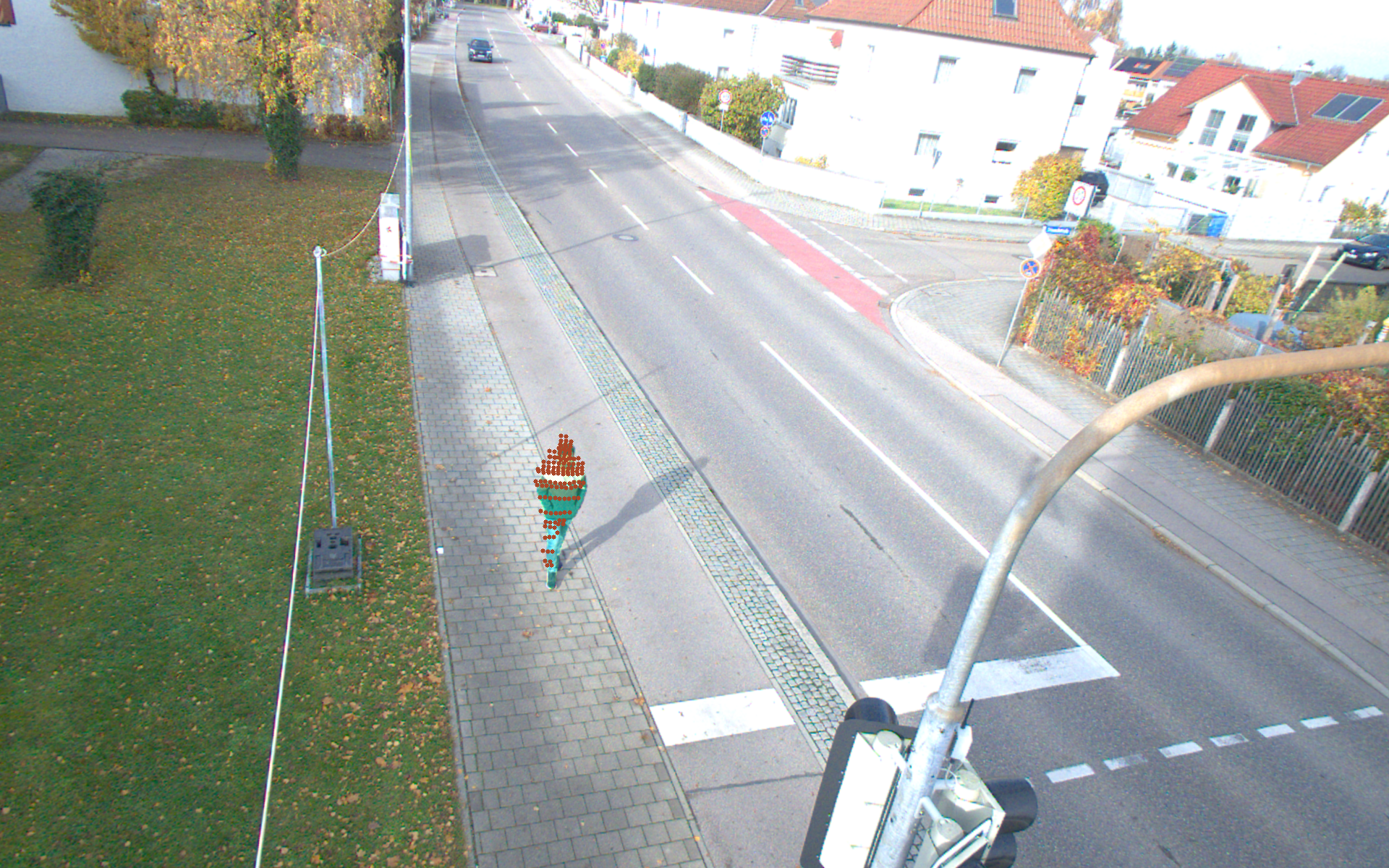}}
    \caption{Person in right camera}
    \label{subfig:camera_3_person}
  \end{subfigure}
  
  \caption{Output of the lidar points projected on the respective camera mask to show the data alignment consistency for all three road users and for all three camera views. The output is generated from the best time-synchronization delay between each camera and lidar frames. These are $30$, $50$, and $65$ milliseconds for left, middle, and right camera with respect to lidar. Fig. (a) to (c) highlights data alignment in the left camera, Fig. (d) to (f) in the middle camera, and Fig. (g) to (i) in the right camera. In every image, only the road user under evaluation is highlighted. }
  \label{fig:time_sync_output_images}
\end{figure*}

Tables~\ref{table:left_camera_coverage_eval},~\ref{table:middle_camera_coverage_eval}, and~\ref{table:right_camera_coverage_eval} show the results of the data alignment for left, middle, and right cameras, respectively. For each camera, three different time-delay configurations are evaluated. These time-delay values are at first obtained by changing the time-delay for a range of values and visually observing the data alignment, as well as using some intuitive understanding of the scan time of lidar with respect to the mounting of each camera. Further, Fig.~\ref{fig:time_sync_output_images} shows the output in the form of projected lidar points on the respective camera object mask for all three cameras using the best time-synchronization delay obtained after evaluations. 

\def\arraystretch{1.2} \tabcolsep=3.5pt
\begin{table}[!h]
\centering
\caption{Left camera average coverage ($\%$) at different distances}
\label{table:left_camera_coverage_eval}
\begin{tabular}{l c c c c}
\toprule \midrule
\multicolumn{1}{c}{\textbf{Time delay}} & \multicolumn{1}{c}{\textbf{distance range (m)}} & \multicolumn{1}{c}{\textbf{Person}} & \multicolumn{1}{c}{\textbf{Bicycle}} & \multicolumn{1}{c}{\textbf{Car}} \\ 
\midrule \hline

\multirow{3}{*}{25 msec} & \multicolumn{1}{c}{Near ($0$ to $15$ m)} & \multicolumn{1}{c}{$76.50$} & \multicolumn{1}{c}{$78.50$} & \multicolumn{1}{c}{$91.88$} \\

& \multicolumn{1}{c}{Mid ($15$ to $30$ m)} & \multicolumn{1}{c}{$65.30$} & \multicolumn{1}{c}{$64.80$} & \multicolumn{1}{c}{$87.84$} \\

& \multicolumn{1}{c}{Far ($30$ to $50$ m)} & \multicolumn{1}{c}{$57.89$} & \multicolumn{1}{c}{$53.14$} & \multicolumn{1}{c}{$81.07$} \\ \hline

\multirow{3}{*}{30 msec} & \multicolumn{1}{c}{Near ($0$ to $15$ m)} & \multicolumn{1}{c}{$82.38$} & \multicolumn{1}{c}{$82.81$} & \multicolumn{1}{c}{$93.18$} \\

& \multicolumn{1}{c}{Mid ($15$ to $30$ m)} & \multicolumn{1}{c}{$66.20$} & \multicolumn{1}{c}{$64.46$} & \multicolumn{1}{c}{$87.81$} \\

& \multicolumn{1}{c}{Far ($30$ to $50$ m)} & \multicolumn{1}{c}{$55.55$} & \multicolumn{1}{c}{$55.57$} & \multicolumn{1}{c}{$76.93$} \\ \hline

\multirow{3}{*}{35 msec} & \multicolumn{1}{c}{Near ($0$ to $15$ m)} & \multicolumn{1}{c}{$79.04$} & \multicolumn{1}{c}{$82.05$} & \multicolumn{1}{c}{$92.53$} \\

& \multicolumn{1}{c}{Mid ($15$ to $30$ m)} & \multicolumn{1}{c}{$56.50$} & \multicolumn{1}{c}{$64.07$} & \multicolumn{1}{c}{$84.34$} \\

& \multicolumn{1}{c}{Far ($30$ to $50$ m)} & \multicolumn{1}{c}{$25.00$} & \multicolumn{1}{c}{$34.61$} & \multicolumn{1}{c}{$78.93$} \\ 

\midrule
\bottomrule
\end{tabular}
\end{table}

\def\arraystretch{1.2} \tabcolsep=3.5pt
\begin{table}[!h]
\centering
\caption{Middle camera average coverage (\%) at different distances}
\label{table:middle_camera_coverage_eval}
\begin{tabular}{cccc}
\toprule
\textbf{Time delay} & \textbf{Person} & \textbf{Bicycle} & \textbf{Car} \\ 
\midrule \hline

$45$ msec & $88.19$ & $89.87$ & $98.77$ \\

$50$ msec & $89.86$ & $89.55$ & $99.19$ \\

$55$ msec & $86.41$ & $88.33$ & $98.95$ \\

\midrule
\bottomrule
\end{tabular}
\end{table}

\def\arraystretch{1.2} \tabcolsep=3.5pt
\begin{table}[!h]
\centering
\caption{Right camera average coverage ($\%$) at different distances}
\label{table:right_camera_coverage_eval}
\begin{tabular}{l c c c c}
\toprule \midrule
\multicolumn{1}{c}{\textbf{Time delay}} & \multicolumn{1}{c}{\textbf{distance range (m)}} & \multicolumn{1}{c}{\textbf{Person}} & \multicolumn{1}{c}{\textbf{Bicycle}} & \multicolumn{1}{c}{\textbf{Car}} \\ 
\midrule \hline

\multirow{3}{*}{60 msec} & \multicolumn{1}{c}{Near ($0$ to $15$ m)} & \multicolumn{1}{c}{$81.69$} & \multicolumn{1}{c}{$69.60$} & \multicolumn{1}{c}{$93.08$} \\

& \multicolumn{1}{c}{Mid ($15$ to $30$ m)} & \multicolumn{1}{c}{$72.72$} & \multicolumn{1}{c}{$64.56$} & \multicolumn{1}{c}{$94.31$} \\

& \multicolumn{1}{c}{Far ($30$ to $50$ m)} & \multicolumn{1}{c}{$56.59$} & \multicolumn{1}{c}{$60.72$} & \multicolumn{1}{c}{$90.36$} \\ \hline

\multirow{3}{*}{65 msec} & \multicolumn{1}{c}{Near ($0$ to $15$ m)} & \multicolumn{1}{c}{$81.93$} & \multicolumn{1}{c}{$77.48$} & \multicolumn{1}{c}{$95.32$} \\

& \multicolumn{1}{c}{Mid ($15$ to $30$ m)} & \multicolumn{1}{c}{$70.94$} & \multicolumn{1}{c}{$66.65$} & \multicolumn{1}{c}{$94.58$} \\

& \multicolumn{1}{c}{Far ($30$ to $50$ m)} & \multicolumn{1}{c}{$65.00$} & \multicolumn{1}{c}{$58.20$} & \multicolumn{1}{c}{$88.41$} \\ \hline

\multirow{3}{*}{70 msec} & \multicolumn{1}{c}{Near ($0$ to $15$ m)} & \multicolumn{1}{c}{$87.79$} & \multicolumn{1}{c}{$72.90$} & \multicolumn{1}{c}{$95.16$} \\

& \multicolumn{1}{c}{Mid ($15$ to $30$ m)} & \multicolumn{1}{c}{$68.14$} & \multicolumn{1}{c}{$59.06$} & \multicolumn{1}{c}{$91.78$} \\

& \multicolumn{1}{c}{Far ($30$ to $50$ m)} & \multicolumn{1}{c}{$49.57$} & \multicolumn{1}{c}{$48.33$} & \multicolumn{1}{c}{$88.76$} \\ 

\midrule
\bottomrule
\end{tabular}
\end{table}

From Table~\ref{table:left_camera_coverage_eval}, it is evident that the coverage of the left camera and corresponding lidar frames is higher for the time delay of $30$ milliseconds. At this time delay, the near distance coverage for person, bicycle, and car are $82.38\%$, $82.82\%$, and $93.18\%$ respectively, which is higher than the respective coverage obtained for time delays of $25$ and $35$ milliseconds. The far range coverage of person and car is slightly higher at a time delay of $25$ milliseconds as compared to $30$ milliseconds, but if considered overall evaluation results, then time delay of $30$ milliseconds suits the best for this camera. Similarly, in Table~\ref{table:right_camera_coverage_eval}, the coverage evaluations indicate that the best data alignment between camera and lidar objects is obtained with a time delay of $65$ milliseconds.

Further, Table~\ref{table:middle_camera_coverage_eval} highlights the evaluations of coverage for middle camera. It is important to note here that the middle camera is mounted upright down (refer to Fig.~\ref{subfig:camera_2_car},~\ref{subfig:camera_2_bicycle}, and~\ref{subfig:camera_2_person} for reference to understand the field-of-view of this camera). As a result, it covers only very small section of the road, particularly the pedestrian crossing. Hence, there is no separate distance ranges possible to evaluate the coverage. Only one value is calculated per road user by considering entire field of view. Table~\ref{table:middle_camera_coverage_eval} shows that this camera achieves relatively high coverage overall, with minimal differences between the three time-delay values tested. Nevertheless, the $50$\,ms time delay provides the maximum coverage across all road users, yielding the best lidar-camera data alignment for the middle camera.

\begin{figure*}[!ht]
\centering
\includegraphics[width=\textwidth]{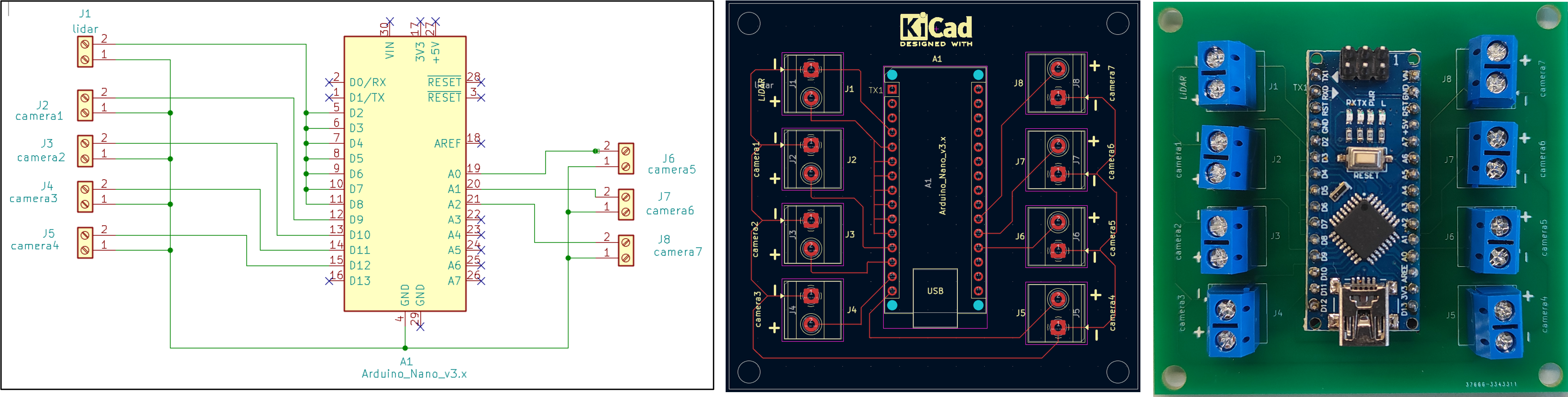}
\caption{Hardware trigger circuit for time synchronization of lidar with seven cameras for vehicle platform: (left) schematic design, (center) KiCad-generated PCB layout, and (right) fabricated and tested PCB.} 
\label{fig:seven_camera_complete_design}
\end{figure*}

Following the successful design, deployment, and evaluation of the hardware-trigger circuit with three cameras for the roadside multi-sensor setup, the same circuit is extended to enable hardware-triggered time synchronization for a sensor car equipped with one Ouster lidar and $7$ Basler cameras of identical model. The schematic design, PCB layout, and fabricated board are shown in Fig.~\ref{fig:seven_camera_complete_design}. This extended circuit is successfully tested in sensor car for its functionality, but detail evaluations are outside the scope of this article and may be considered in the future work.

Further, regarding generalization, although the current implementation leverages Ouster-specific functionality, the underlying architecture defines a vendor-agnostic, hardware-triggered logic chain. This synchronization framework is directly extensible to any lidar sensor supporting physical Phase Lock and Trigger Out signals, such as the Hesai Pandar64 \cite{Hesai2025Pandar64}, RoboSense RS-32 \cite{robosense_rs_32_user_guide}, etc. Similarly, many industrial-grade cameras, such as IDS \cite{ids} and Teledyne FLIR \cite{flir_teledyne_vision_solutions}, etc., feature dedicated GPIO trigger inputs as a standard, facilitating broad compatibility with this hardware-centric approach.

Extending this hardware-centric approach to broader infrastructure, the system may be scaled across distributed infrastructure, such as multiple sensor masts deployed at an intersection. In such a multi-node deployment, the same hardware logic can be replicated at each mast to maintain local camera-lidar time synchronization. To achieve global synchronization, each node should then be interfaced with a common PTP \cite{ieee_ptp} Grandmaster, which can be (optionally) further referenced to a GPS clock as the absolute time source. By utilizing such a PTP backbone, each distributed lidar can be phase-locked to a shared global reference, ensuring a deterministic firing sequence across the network. This approach can ensure that data from disparate nodes is natively aligned at the hardware level. Consequently, this may eliminate the need for complex temporal compensation at the Cooperative Perception Service (CPS) level, facilitating seamless, real-time data fusion.
\section{Conclusion} \label{sec:conclusion}

In this work, a simple, modular, open-source, and configurable hardware-triggered time-synchronization circuit is designed and validated to achieve precise synchronization between a lidar and three cameras in a roadside-mounted multi-sensor system deployed near a school. The circuit utilizes the lidar synchronization pulse as a reference input and generates independently programmable, time-delayed trigger pulses for each camera. This design enables flexible adaptation to different sensor configurations and mounting geometries. Comprehensive experiments were conducted to determine, both statistically and visually, the optimal time delays for each camera relative to the lidar scan to achieve the highest temporal alignment of road users across modalities. The results indicate that the best time delays are $30$, $50$, and $65$ milliseconds for the left, middle, and right cameras, respectively, relative to the lidar. Furthermore, the circuit design is extended to support hardware-triggered time synchronization in a sensor- equipped vehicle integrating one Ouster lidar and seven Basler cameras of the same model.


\bibliography{references}

@inproceedings{liGlobalClockSynchronization2004,
  title = {Global Clock Synchronization in Sensor Networks},
  booktitle = {{{IEEE INFOCOM}} 2004},
  author = {Li, Qun and Rus, D.},
  year = {2004},
  month = {mar},
  volume = {1},
  pages = {574},
  issn = {0743-166X},
  doi = {10.1109/INFCOM.2004.1354528},
  urldate = {2025-10-07}
}

@inproceedings{nodaFrameSynchronizationNetworked2014,
  title = {Frame Synchronization for Networked High-Speed Vision Systems},
  booktitle = {2014 {{IEEE SENSORS}}},
  author = {Noda, Akihito and Yamakawa, Yuji and Ishikawa, Masatoshi},
  year = {2014},
  month = {nov},
  pages = {269--272},
  issn = {1930-0395},
  doi = {10.1109/ICSENS.2014.6984985},
  urldate = {2025-10-07}
}

@inproceedings{englishTriggerSyncTimeSynchronisation2015,
  title = {{{TriggerSync}}: {{A}} Time Synchronisation Tool},
  shorttitle = {{{TriggerSync}}},
  booktitle = {2015 {{IEEE International Conference}} on {{Robotics}} and {{Automation}} ({{ICRA}})},
  author = {English, Andrew and Ross, Patrick and Ball, David and Upcroft, Ben and Corke, Peter},
  year = {2015},
  month = {may},
  pages = {6220--6226},
  issn = {1050-4729},
  doi = {10.1109/ICRA.2015.7140072},
  urldate = {2025-10-07}
}

@article{hylaMultiCameraTriggering2016,
  title = {Multi Camera Triggering and Synchronization Issue : Case Study},
  shorttitle = {Multi Camera Triggering and Synchronization Issue},
  author = {Hyla, P.},
  year = {2016},
  journal = {Journal of KONES},
  urldate = {2025-10-07},
  langid = {english}
}

@inproceedings{sommerLowcostSystemHighrate2017,
  title = {A Low-Cost System for High-Rate, High-Accuracy Temporal Calibration for {{LIDARs}} and Cameras},
  booktitle = {2017 {{IEEE}}/{{RSJ International Conference}} on {{Intelligent Robots}} and {{Systems}} ({{IROS}})},
  author = {Sommer, Hannes and Khanna, Raghav and Gilitschenski, Igor and Taylor, Zachary and Siegwart, Roland and Nieto, Juan},
  year = {2017},
  month = {sep},
  pages = {2219--2226},
  issn = {2153-0866},
  doi = {10.1109/IROS.2017.8206042},
  urldate = {2025-10-07}
}

@inproceedings{huSoftTimeSynchronization2018,
  title = {A {{Soft Time Synchronization Framework}} for {{Multi-Sensors}} in {{Autonomous Localization}} and {{Navigation}}},
  booktitle = {2018 {{IEEE}}/{{ASME International Conference}} on {{Advanced Intelligent Mechatronics}} ({{AIM}})},
  author = {Hu, hang and Wu, Jianhua and Xiong, Zhenhua},
  year = {2018},
  month = {jul},
  pages = {694--699},
  issn = {2159-6255},
  doi = {10.1109/AIM.2018.8452384},
  urldate = {2025-10-07}
}

@inproceedings{osadcuksClockbasedTimeSynchronization2020,
  title = {Clock-Based Time Synchronization for an Event-Based Camera Dataset Acquisition Platform},
  booktitle = {2020 {{IEEE International Conference}} on {{Robotics}} and {{Automation}} ({{ICRA}})},
  author = {Osadcuks, Vitalijs and Pudzs, Mihails and Zujevs, Andrejs and Pecka, Aldis and Ardavs, Arturs},
  year = {2020},
  month = {may},
  pages = {4695--4701},
  issn = {2577-087X},
  doi = {10.1109/ICRA40945.2020.9197303},
  urldate = {2025-10-07}
}

@inproceedings{liuBriefIndustryPaper2021,
  title = {Brief {{Industry Paper}}: {{The Matter}} of {{Time}} --- {{A General}} and {{Efficient System}} for {{Precise Sensor Synchronization}} in {{Robotic Computing}}},
  shorttitle = {Brief {{Industry Paper}}},
  booktitle = {2021 {{IEEE}} 27th {{Real-Time}} and {{Embedded Technology}} and {{Applications Symposium}} ({{RTAS}})},
  author = {Liu, Shaoshan and Yu, Bo and Liu, Yahui and Zhang, Kunai and Qiao, Yisong and Li, Thomas Yuang and Tang, Jie and Zhu, Yuhao},
  year = {2021},
  month = {may},
  pages = {413--416},
  issn = {2642-7346},
  doi = {10.1109/RTAS52030.2021.00040},
  urldate = {2025-10-07}
}

@article{grammatikopoulosEffectiveCameratoLidarSpatiotemporal2022,
  title = {An {{Effective Camera-to-Lidar Spatiotemporal Calibration Based}} on a {{Simple Calibration Target}}},
  author = {Grammatikopoulos, Lazaros and Papanagnou, Anastasios and Venianakis, Antonios and Kalisperakis, Ilias and Stentoumis, Christos},
  year = {2022},
  month = {jan},
  journal = {Sensors},
  volume = {22},
  number = {15},
  pages = {5576},
  publisher = {Multidisciplinary Digital Publishing Institute},
  issn = {1424-8220},
  doi = {10.3390/s22155576},
  urldate = {2025-10-07},
  copyright = {http://creativecommons.org/licenses/by/3.0/},
  langid = {english}
}

@article{yuanLiCaS3SimpleLiDAR2022,
  title = {{{LiCaS3}}: {{A Simple LiDAR}}--{{Camera Self-Supervised Synchronization Method}}},
  shorttitle = {{{LiCaS3}}},
  author = {Yuan, Kaiwen and Ding, Li and Abdelfattah, Mazen and Wang, Z. Jane},
  year = {2022},
  month = {oct},
  journal = {IEEE Transactions on Robotics},
  volume = {38},
  number = {5},
  pages = {3203--3218},
  issn = {1941-0468},
  doi = {10.1109/TRO.2022.3167455},
  urldate = {2025-10-07}
}

@article{wangTimeSynchronizationSpace2023,
  title = {Time {{Synchronization}} and {{Space Registration}} of {{Roadside LiDAR}} and {{Camera}}},
  author = {Wang, Chuan and Liu, Shijie and Wang, Xiaoyan and Lan, Xiaowei},
  year = {2023},
  month = {jan},
  journal = {Electronics},
  volume = {12},
  number = {3},
  pages = {537},
  publisher = {Multidisciplinary Digital Publishing Institute},
  issn = {2079-9292},
  doi = {10.3390/electronics12030537},
  urldate = {2025-10-07},
  copyright = {http://creativecommons.org/licenses/by/3.0/},
  langid = {english}
}

@inproceedings{kuhseSyncSinkRobustness2024,
  title = {Sync or {{Sink}}? {{The Robustness}} of {{Sensor Fusion Against Temporal Misalignment}}},
  shorttitle = {Sync or {{Sink}}?},
  booktitle = {2024 {{IEEE}} 30th {{Real-Time}} and {{Embedded Technology}} and {{Applications Symposium}} ({{RTAS}})},
  author = {Kuhse, Daniel and Holscher, Nils and Gunzel, Mario and Teper, Harun and Von Der Bruggen, Georg and Chen, Jian-Jia and Lin, Ching-Chi},
  year = {2024},
  month = {may},
  pages = {122--134},
  issn = {2642-7346},
  doi = {10.1109/RTAS61025.2024.00018},
  urldate = {2025-10-07}
}

@inproceedings{wangHardwareBasedTimeSynchronization2024,
  title = {Hardware-{{Based Time Synchronization}} for a {{Multi-Sensor System}}},
  booktitle = {2024 {{IEEE}}/{{RSJ International Conference}} on {{Intelligent Robots}} and {{Systems}} ({{IROS}})},
  author = {Wang, Yueqi and Liu, Tangyou and Feng, Licheng and Wang, Jinze and Yang, Yang and Bao, Jianjun and Li, Binghao and Wu, Liao},
  year = {2024},
  month = {oct},
  pages = {4600--4607},
  issn = {2153-0866},
  doi = {10.1109/IROS58592.2024.10802693},
  urldate = {2025-10-07}
}

@article{gongRoadsideLiDARCameraFusion2025,
  title = {Roadside {{LiDAR-Camera Fusion Detection Based}} on {{Spatiotemporal Calibration}}},
  author = {Gong, Bowen and Wang, Yimeng and Lin, Ciyun and Liu, Hongchao},
  year = {2025},
  month = {aug},
  journal = {IEEE Sensors Journal},
  volume = {25},
  number = {16},
  pages = {31313--31325},
  issn = {1558-1748},
  doi = {10.1109/JSEN.2025.3588678},
  urldate = {2025-10-07}
}

@article{daiLiDARCameraSpatiotemporal2025,
  title = {A {{LiDAR}}--{{Camera Spatiotemporal Synchronization Method}} for {{Unmanned Aerial Vehicle-Based Ground Target Perception}}},
  author = {Dai, Keren and Zhu, Jie and Hou, Haiting and Li, Qingyu and Ma, Xiang and Yu, Hang},
  year = {2025},
  journal = {IEEE Transactions on Instrumentation and Measurement},
  volume = {74},
  pages = {1--15},
  issn = {1557-9662},
  doi = {10.1109/TIM.2025.3558243},
  urldate = {2025-10-07}
}

@article{gurumadaiahPreciseSynchronizationLiDAR2025,
  title = {Precise {{Synchronization Between LiDAR}} and {{Multiple Cameras}} for {{Autonomous Driving}}: {{An Adaptive Approach}}},
  shorttitle = {Precise {{Synchronization Between LiDAR}} and {{Multiple Cameras}} for {{Autonomous Driving}}},
  author = {Gurumadaiah, Ajay Kumar and Park, Jaehyeong and Lee, Jin-Hee and Kim, JeSeok and Kwon, Soon},
  year = {2025},
  month = {mar},
  journal = {IEEE Transactions on Intelligent Vehicles},
  volume = {10},
  number = {3},
  pages = {2152--2162},
  issn = {2379-8904},
  doi = {10.1109/TIV.2024.3444780},
  urldate = {2025-10-07}
}

@ARTICLE{shiva_auto_calibration_paper,
  author={Agrawal, Shiva and Bhanderi, Savankumar and Doycheva, Kristina and Elger, Gordon},
  journal={IEEE Sensors Journal}, 
  title={Static Multitarget-Based Autocalibration of RGB Cameras, 3-D Radar, and 3-D Lidar Sensors}, 
  year={2023},
  volume={23},
  number={18},
  pages={21493-21505},
  doi={10.1109/JSEN.2023.3300957}
}

@STANDARD{ieee_ptp,
  title={IEEE Standard for a Precision Clock Synchronization Protocol for Networked Measurement and Control Systems}, 
  organization={IEEE},
  number={IEEE Std 1588-2019 (Revision of IEEE Std 1588-2008)},
  year={2020},
  pages={1-499},
  doi={10.1109/IEEESTD.2020.9120376}}

@ARTICLE{ieee_ntp,
  author={Mills, D.L.},
  journal={IEEE Transactions on Communications}, 
  title={Internet time synchronization: the network time protocol}, 
  year={1991},
  volume={39},
  number={10},
  pages={1482-1493},
  doi={10.1109/26.103043}}

@Article{time_delay_measurement,
AUTHOR = {Yusupov, Anvarjon and Park, Sun and Kim, JongWon},
TITLE = {Synchronized Delay Measurement of Multi-Stream Analysis over Data Concentrator Units},
JOURNAL = {Electronics},
VOLUME = {14},
YEAR = {2025},
NUMBER = {1},
ARTICLE-NUMBER = {81},
URL = {https://www.mdpi.com/2079-9292/14/1/81},
ISSN = {2079-9292},
DOI = {10.3390/electronics14010081}
}

@Article{time_sync_basics,
AUTHOR = {Yeong, De Jong and Velasco-Hernandez, Gustavo and Barry, John and Walsh, Joseph},
TITLE = {Sensor and Sensor Fusion Technology in Autonomous Vehicles: A Review},
JOURNAL = {Sensors},
VOLUME = {21},
YEAR = {2021},
NUMBER = {6},
ARTICLE-NUMBER = {2140},
URL = {https://www.mdpi.com/1424-8220/21/6/2140},
PubMedID = {33803889},
ISSN = {1424-8220},
DOI = {10.3390/s21062140}
}

@InProceedings{CCIS_springer_chapter_2023,
  author    = {Shiva Agrawal and Rui Song and Krasimira Doycheva and Alois Knoll and Gordon Elger},
  title     = {Intelligent Roadside Infrastructure for Connected Mobility},
  booktitle = {Smart Cities, Green Technologies, and Intelligent Transport Systems},
  editor    = {Cornel Klein and Matthias Jarke and Jeroen Ploeg and Martin Helfert and Karsten Berns and Oleg Gusikhin},
  series    = {Communications in Computer and Information Science},
  volume    = {1843},
  pages     = {80--99},
  year      = {2023},
  publisher = {Springer, Cham},
  doi       = {10.1007/978-3-031-37470-8_6},
  label = {Agr3}
}

@Article{mdpi_sustainability_mmwave_camera_fusion_review,
AUTHOR = {Zhou, Yong and Dong, Yanyan and Hou, Fujin and Wu, Jianqing},
TITLE = {Review on Millimeter-Wave Radar and Camera Fusion Technology},
JOURNAL = {Sustainability},
VOLUME = {14},
YEAR = {2022},
NUMBER = {9},
ARTICLE-NUMBER = {5114},
URL = {https://www.mdpi.com/2071-1050/14/9/5114},
ISSN = {2071-1050},
DOI = {10.3390/su14095114}
}

@misc{ousterOS1_datasheet,
author= {{Ouster}},
  title = {Ouster OS1 lidar sensor Datasheet},
  url = {https://data.ouster.io/downloads/datasheets/datasheet-rev06-v2p4-os1.pdf},
  urldate = {2025-12-29}
}

@misc{ouster_osdome_datasheet,
  author   = {{{Ouster}}},
  title    = {Ouster OSdome lidar sensor Datasheet},
  url      = {https://data.ouster.io/downloads/datasheets/datasheet-rev7-v3p0-osdome.pdf},
  urldate  = {2025-12-29}
}

@misc{basler_gige_camera,
  author = {{Basler AG}},
  title = {Basler ace 2 a2A1920-51gcBAS camera},
  url = {https://www.baslerweb.com/en/shop/a2a1920-51gcbas},
  urldate = {2025-12-29}
}

@misc{ros_pc2_msg,
  author = {{ROS}},
  title = {ROS point cloud 2 message description},
  url = {http://docs.ros.org/en/noetic/api/sensor_msgs/html/msg/PointCloud2.html},
  urldate = {2025-12-29}
}

@misc{ros_image_msg,
   author = {{ROS}},
  title = {ROS Image message description},
  url = {https://docs.ros.org/en/noetic/api/sensor_msgs/html/msg/Image.html},
  urldate = {2025-12-29}
}

@article{ros2,
    author = {Steven Macenski and Tully Foote and Brian Gerkey and Chris Lalancette and William Woodall},
    title = {Robot Operating System 2: Design, architecture, and uses in the wild},
    journal = {Science Robotics},
    volume = {7},
    number = {66},
    pages = {eabm6074},
    year = {2022},
    doi = {10.1126/scirobotics.abm6074},
    URL = {https://www.science.org/doi/abs/10.1126/scirobotics.abm6074}
}

@techreport{ouster_os1_hardware_user_manual_rev7,
author= {{Ouster}},
  title = {OS1 Hardware User Manual (Rev7)},
  institution = {Ouster},
  year = {2023},
  url = {https://data.ouster.io/downloads/hardware-user-manual/hardware-user-manual-rev7-os1.pdf},
  note = {Accessed: Dec. 31, 2025}
}

@techreport{ouster_interface_box,
  author= {{{Ouster}}},
  title = {OS1 Electrical Interface Documentation},
  institution = {Ouster},
  year = {2023},
  url = {https://static.ouster.dev/sensor-docs/hw_user_manual_OS1/hw_common_sections_OS1/Accessories.html},
  urldate = {2025-12-31}
}

@inproceedings{mask_rcnn,
  title={Mask r-cnn},
  author={He, Kaiming and Gkioxari, Georgia and Doll{\'a}r, Piotr and Girshick, Ross},
  booktitle={Proceedings of the IEEE international conference on computer vision},
  pages={2961--2969},
  year={2017}
}

@INPROCEEDINGS{sushtech_points_labeling_tool,
  author={Li, E and Wang, Shuaijun and Li, Chengyang and Li, Dachuan and Wu, Xiangbin and Hao, Qi},
  booktitle={2020 IEEE Intelligent Vehicles Symposium (IV)}, 
  title={SUSTech POINTS: A Portable 3D Point Cloud Interactive Annotation Platform System}, 
  year={2020},
  volume={},
  number={},
  pages={1108-1115},
  doi={10.1109/IV47402.2020.9304562}}

@misc{flir_teledyne_vision_solutions,
   author = {{Teledyne Vision Solutions}},
  title = {Configuring Synchronized Capture with Multiple Cameras},
  url = {https://www.teledynevisionsolutions.com/support/support-center/application-note/iis/configuring-synchronized-capture-with-multiple-cameras/},
  urldate = {2026-05-04}
}

@misc{ids,
   author = {{iDS}},
  title = {Synchronous image acquisition},
  url = {https://en.ids-imaging.com/techtipp-details/items/app-note-synchronizing-image-acquisition.html},
  urldate = {2026-05-04}
}

@manual{robosense_rs_32_user_guide,
  title        = {RS-LiDAR-32 User Guide},
  author       = {{Robosense}},
  organization = {RoboSense},
  url          = {https://www.mybotshop.de/Datasheet/Robosense_RS-32_User_Guide.pdf},
  urldate = {2026-05-04}
}

@manual{Hesai2025Pandar64,
  title        = {Pandar64 64-Channel Mechanical Lidar User Manual},
  author       = {{Hesai Technology Co., Ltd.}},
  organization = {Hesai Technology},
  note         = {Document Version: 640-en-250410},
  url          = {https://www.hesaitech.com/wp-content/uploads/2025/04/Pandar64_User_Manual_640-en-250410.pdf},
  urldate = {2026-05-04}
}
\bibliographystyle{IEEEtran}

\end{document}